\documentclass[journal,twoside,web]{ieeecolor}
\usepackage[ruled,vlined]{algorithm2e}

\usepackage{jsen}
\usepackage{cite}
\usepackage{algorithmic}
\usepackage{graphicx}
\usepackage{wrapfig}
\usepackage{multirow}

\def\BibTeX{{\rm B\kern-.05em{\sc i\kern-.025em b}\kern-.08em
    T\kern-.1667em\lower.7ex\hbox{E}\kern-.125emX}}
\definecolor{abstractbg}{rgb}{1,1,1}

\begin{document}
\title{Multimedia Datasets for Anomaly Detection: A Review}
\author{Pratibha Kumari, Anterpreet Kaur Bedi and Mukesh Saini, \IEEEmembership{Indian Institute of Technology Ropar, India}
%\thanks{This paragraph of the first footnote will contain the date on which you submitted your paper for review. It will also contain support information, including sponsor and financial support acknowledgment. For example, ``This work was supported in part by the U.S. Department of 
%Commerce under Grant BS123456.'' }
%\thanks{The next few paragraphs should contain 
%the authors' current affiliations, including current address and e-mail. For 
% example, F. A. Author is with the National Institute of Standards and 
% Technology, Boulder, CO 80305 USA (e-mail: author@boulder.nist.gov). }
% \thanks{S. B. Author, Jr., was with Rice University, Houston, TX 77005 USA. He is 
% now with the Department of Physics, Colorado State University, Fort Collins, 
% CO 80523 USA (e-mail: author@lamar.colostate.edu).}
% \thanks{T. C. Author is with 
% the Electrical Engineering Department, University of Colorado, Boulder, CO 
% 80309 USA, on leave from the National Research Institute for Metals, 
% Tsukuba, Japan (e-mail: author@nrim.go.jp).}
}
\IEEEtitleabstractindextext{%
\fcolorbox{abstractbg}{abstractbg}{%
 \begin{minipage}{\textwidth}%
%  \begin{wrapfigure}[12]{r}{3in}%
%  \includegraphics[width=3in]{jsenga.png}%
%  \end{wrapfigure}%

\begin{abstract}
Multimedia anomaly datasets play a crucial role in automated surveillance. They have a wide range of applications expanding from outlier objects/ situation detection to the detection of life-threatening events. For more than 1.5 decades, this field has attracted a lot of research attention, and as a result, more and more datasets dedicated to anomalous actions and object detection have been developed. Tapping these public anomaly datasets enable researchers to generate and compare various anomaly detection frameworks with the same input data. This paper presents a comprehensive survey on a variety of video, audio, as well as audio-visual datasets based on the application of anomaly detection. This survey aims to address the lack of a comprehensive comparison and analysis of multimedia public datasets based on anomaly detection. Also, it can assist researchers in selecting the best available dataset for bench-marking frameworks. Additionally, we discuss gaps in the existing dataset and insights for future direction towards developing multimodal anomaly detection datasets.
\end{abstract}
\begin{IEEEkeywords}
Anomaly datasets survey, Concept drift, Multimodal anomaly detection, Long term surveillance
\end{IEEEkeywords}
\end{minipage}
}}
\maketitle
\section{Introduction}
%-------------------------------
\begin{figure*}
  \centering
    \includegraphics[scale=0.55]{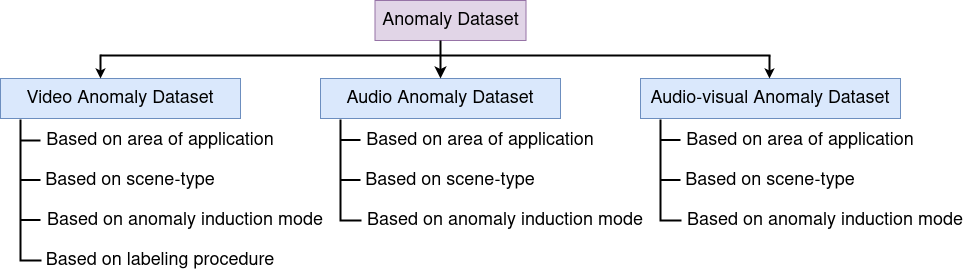}
  \caption{Taxonomy for multimedia anomaly datasets}
  \label{fig:taxonomy}
\end{figure*}
%-------------------------------
Anomaly detection corresponds to identifying unexpected or rare events in a dataset~\cite{chandola2009anomaly}. Detection of anomalous events from a scenario can offer important insights into a large number of monitoring and safety-critical real-world applications such as disaster forecast, detection of extreme climate event, mechanical fault, disease outbreak, fire/ blast, fraud, etc. 
%This field has been studied in a number of research applications owing to its broad applicability. Additionally, due to easy availability of economic microphones, CCTV cameras and fast audio/ video processing power, applications relying on these modalities have seen a tremendous growth. 
Recently, there has been a significant rise in the use of audio and video sensors for monitoring situations~\cite{jaafar2019audio,rehman2021multi,kumari2021situational}. Monitoring of scenes by humans is unscalable and error-prone due to inherent human limitations~\cite{elharrouss2021review,kumari2020multivariate}. Therefore, we need to automatically analyse audio/video data to detect anomalous events and objects. Consequently, many frameworks for anomaly detection have been proposed. In order to evaluate and validate a framework, researchers require a comprehensive dataset. Towards this end, numerous datasets have been created in literature for anomaly detection. These are either based on a specific type of anomaly, or for generic anomaly detection.

In this survey paper, we present a comprehensive review on video, audio, and audiovisual datasets available for evaluating anomaly detection frameworks. We compare these datasets based on various perspectives, which will help researchers to choose suitable datasets for an application. To the best of our knowledge, there exists only one short survey paper~\cite{patil2016survey} on video datasets published in 2016. We could not find any survey on audio or audio-visual anomaly datasets. We believe that a consolidated review of audio, video, and audio-visual datasets will promote use of multimedia for anomaly detection. Additionally, there are more than fifteen new datasets contributed to the video anomaly detection domain after the year 2016, which need to be reviewed.

%However, we present video datasets from diverse applications, audio datasets, audio-visual datasets in a structured manner. Additionally, there are more than fifteen new datasets contributed to the video anomaly detection domain after the year 2016, which generates the need to explain and compare those datasets. 

%%Further, they can be categorized on different attribute, e.g., based on way of labeling, application, scene-type, etc.
A comprehensive taxonomy of anomaly datasets is given in Figure~\ref{fig:taxonomy}. Based on the recording media, anomaly datasets can be categorized into video, audio, and audio-visual datasets. 
Further, we categorize them on the basis of four attributes, viz., area of application, scene-type, anomaly induction mode, and labeling type. 
%Such categorization and comparison on common attributes aid in their comprehension. %%% and will be helpful in selecting datasets according to specific requirement of surveillance. 
%%Datasets are compared and discussed in detail based on these attributes in next sections.
%%They are described on a common set of attributes to aid in their comprehension. 
The characterization is expanded to include additional attributes in multiple figures and tables across Sections~\ref{sec:videoDatasets}, \ref{sec:audioDatasets}, and \ref{sec:audioVisual}. 
This enables a quick comparison of existing anomaly datasets, and thus helpful in selecting most suitable dataset to evaluate the given work.
Further, to facilitate an easy access of datasets, we assemble them together and provide brief description including common information such as how and where the data was collected, what are the anomalies, merit/ demerit identified by researchers, etc., along with link to their website via supplementary file attached with this manuscript.

Rest of the paper is organized as follows. Public video datasets for anomaly detection are discussed in the Section~\ref{sec:videoDatasets}. A discussion about the audio datasets has been provided in Section~\ref{sec:audioDatasets}. Further, a comparison of public audio-visual datasets is given in Section~\ref{sec:audioVisual}. Section~\ref{sec:futureDirection} discusses on future directions. Finally, we conclude the paper in Section~\ref{sec:conclusions}.
%%%%%%%%%%%%%%%%%%%%%%%%%%%%%%%%%%%%%%%%%%%%%%%%%%%%%%%%%%%%%%%
%%%%%%%%%%%%%%%%%%%%%%%%%%%%%%%%%%%%%%%%%%%%%%%%%%%%%%%
\section{Video anomaly datasets}\label{sec:videoDatasets}
There are more than 30 publicly available video datasets that are currently being used for the purpose of anomaly detection. 
Initial datasets are comprised of simple events and scenarios with a very constrained amount of anomalies. The anomalies are performed by a group of actors and therefore lack the natural flow of events. They are of very short duration as well. The presence of a few rare objects or events in unimodal background events was regarded as anomaly in these datasets. Some examples of such datasets are Canoe~\cite{jodoin2008modeling} (a boat occurring once in the scene is regarded anomaly), UMN~\cite{xyz} (few people acting for sudden evacuation is regarded anomaly here), Web~\cite{xyz} (panic-escape and crowd fighting regarded as anomaly), Subway entrance/exit~\cite{adam2008robust} (movement in wrong direction regarded as anomaly), various sub-clips in abnormal behavior dataset (appearance of only one type of rare object/event), etc. 

Later datasets consist of more types of anomalies and scene variability such as UCSD~\cite{mahadevan2010anomaly}, AVENUE~\cite{lu2013abnormal}, ARENA~\cite{pets2014}, ShanghaiTech~\cite{luo2017revisit}. In contrast, some recently developed datasets are gigantic in terms of duration and variability of scene and events such as UCF-Crime~\cite{sultani2018real}, VIRAT~\cite{oh2011large}, LV~\cite{leyva2017lv}, ADOC~\cite{pranav2020day}, HTA~\cite{singh2020anomalous}, Rodriguez's~\cite{rodriguez2011data}, etc. They leverage the online sources of CCTV footage and public videos and curate large databases of realistic video clips from our daily life. They are equipped with a variety of events and actions in with complex scene settings. Some datasets are suitable for generic scene monitoring, such as UCSD~\cite{mahadevan2010anomaly}, ADOC~\cite{pranav2020day}, AVENUE~\cite{lu2013abnormal}, ShanghaiTech~\cite{luo2017revisit}, etc., while others are suitable for specific classes of anomaly detection such as Subway entrance/exit~\cite{adam2008robust}, UMN~\cite{xyz}, Web~\cite{xyz}, i-Lids~\cite{iLid}, etc. Further, few datasets such as UCF-Crime, LV, and NVIDIA AI CITY~\cite{nvidia} contain specific anomaly classes that are important for usage in frameworks for detecting life-threatening events.
%=========================================  
%=====================================================
\begin{table*}[!htbp]
  \caption{Specifications of video anomaly datasets}
  \label{tab:videoTableBig4}
   \centering
  \resizebox{.99\textwidth}{!}{
  \begin{tabular}{|c|c|c|c|c|c|c|c|}
    \hline
    
Dataset& Continuity&\begin{tabular}[c]{@{}c@{}} Total\\Duration\end{tabular}   & \begin{tabular}[c]{@{}c@{}} Total No. of\\Frames\end{tabular}   &   \begin{tabular}[c]{@{}c@{}} Total No. of\\Videos\end{tabular} & Resolution&FPS&\begin{tabular}[c]{@{}c@{}} Camera\\motion\end{tabular}  \\\hline
       
i-Lids (2007)&no&$\approx$24 min &35000&7& 720×576&25&none\\
QMUL (2008)&no& 22 min&34000&104(N)+8(A)&360×288 &25&none\\
Canoe (2008)&yes&	34 s &1050&1&320 × 240&30&none\\
Subway Entrance (2008)&yes& 96 m 9 s  &144225 &1&512×384&25&none\\
Subway Exit (2008) & yes&43 m 16 s &64900&1&512×384&25&none \\
UMN (2009) & no&4 min 17 sec & 7710&11 (available as 1)&320×240&30&none\\
Web (2009) &no&  7 min 35 sec& 11,962 &12(N)+8(A)&multiple&multiple&slight jerks\\
PETS2009 (2009) &no&$\approx$ 1-2 hrs &42182&59& \begin{tabular}[c]{@{}c@{}} 768x576 \\720x576\end{tabular}&7&none\\

U-Turn (2009) &no&$\approx$20 min&$\approx$25182&8&360 × 240&multiple&none\\
Idiap (2009) &yes& 44.13 min&66324&1&288×360&25&none\\

USCD Ped1 (2010) &no&$\approx$ 5-7 min  & 14000&34(N)+36(AN)&238×158&-& none\\
UCSD Ped2 (2010)&no&$\approx$ 2-3 min  & 4560&16(N)+14(AN)&360×240&-&none \\
Train (2010)&	yes&12 min&19218&1&288×386&25&moving\\
Belleview (2010)&no&4 min 51 sec	&2918&1&320 × 240&10&jitter\\
Boat-Sea (2010)&yes&1 min 56 sec	&450&1	& 720×576&19&none\\
Boat-River (2010)&yes&1 min 8 sec&	250	&1& 704×576&5&none\\
Caouflage (2010)&	yes&54 sec&1629	&1&320×240&29.97&none\\
Rodriguez's&no&10 hrs 24 min&-&520&720x480&-&none\\
Hockey (2011)&no&27 min&50000&1000& 720×576&-&none\\
Movie (2011)&no&6 min&-&200&multiple&multiple&in some\\
UCF Crowd (2012)&no&$\approx$11 min&$\approx$16320&38&multiple&multiple&none\\
Voilent-Flows (2012)&no&14 min 6 sec&22,156 &246&320×240&multiple&significant\\
Grand Central Station (2012)&yes&33 min 20 sec&50010 &1&480×720&25&none\\
AGORASET (2012)& no&$\approx$20 min & $>$33641&23 &640×480&30&none\\
Meta-tracking (2013)&no&--&$>$4000&12 &multiple&multiple&none\\

Avenue (2013) & no&20 min 26 sec &30652 & 16(N)+21(AN)& 640×360&25&slight in few\\
ARENA (2014)&no&--&--&22& 1280 x 960 &30  &none\\

PWPD (2015)&yes&1 hour&5000&1&1920×1080&1.25&none\\
RE-DID (2015)&no& $<$2 hrs  &  - &30&1280X720&multiple&in some\\
MED (2016)& no&$\approx$24-25 min & 43,626&31&554×235&31&none\\

ShanghaiTech (2017)&no& $\approx$ 3-3.5 hrs  &317398 &330(N)+107(AN)&856×480&24&none \\
LV (2017)&no&3.93 hrs&-&30&multiple&multiple&in some\\
UCF-Crime (2018)&no& $\approx$ 128 hrs &$\approx$ 13 million &950(N)+950(A)&320×240&30&slight in few \\
IITH Accidents (2018)&no&--&128001 &-- & --&30&none\\
CCTV-Fights (2019)&no&$\approx$ 10 hrs & -& 1000& multiple&multiple&in some\\

Street Scene (2020)&no&$\approx$ 3-4 h&203257&46(N)+35(AN)&1280×720&15&none\\
ADOC (2020)&yes&24 hrs&259127&4&2048 × 1536&3&none\\
HTA (2020)& no&$\approx$ 4 hrs &$\approx$ 0.4 million& 286(N)+107(AN)&1280×720&30&moving \\ \hline
\end{tabular}
}
\end{table*}
%===============================================================
%=========================================
%
% \hl{A brief introduction of all the publicly available video anomaly datasets is given through Tables{~\ref{tab:videoTableMain1}} to{~\ref{tab:videoTableMain3}}. The datasets are listed in 
% increasing order of their release year. Some examples of anomalies present and the recording location of the dataset is mentioned in the second column.
% Information about dataset collection scene, i.e., whether it is a mall or an airport or a traffic junction, etc., may help to choose scene-specific dataset for the evaluation of scene-specific surveillance frameworks. The third column shows the area of application which gives an idea of different applications where the dataset has been used for bench-marking. Visuals from the dataset, i.e., a sample normal and an abnormal frame from the dataset is shown via the last two columns.}
%%%%%%%%%%%%%%%%%%%%%%%%%%%%%%%%%%%%%%%%%%%%%%%%%%%%%%%%%%%%%%%%%%%%%%%%%%%%%
%\subsection{Dataset specifications}
%%%\hl{Tables{~\ref{tab:videoTableMain1}} to {\ref{tab:videoTableMain3}} gives an overall picture of all available datasets for video anomaly detection. Once a researcher finalize the application or specific scene; there can be multiple candidate datasets. Selecting the most suitable ones from them require further digging into dataset specifications. Specifications of datasets includes the details of recorded footage, e.g., frame rate, resolution, duration, etc.}

Table~\ref{tab:videoTableBig4} presents a comparison of video anomaly datasets on various attributes of recorded videos such as continuity, duration, number of frames, number of video clips, resolution, frame rate, and camera motion. The datasets are listed in chronological order of their release year. Continuity (yes if untrimmed) information helps to see the applicability of a dataset for the task at hand. If the aim is to examine the context adaptation of anomaly detection framework, then the footage should be untrimmed as well as recorded at a single location~\cite{kumari2020multivariate,kumari2021situational}. The adaptive models need sufficient data to learn the context on their own and adapt over time; hence a dataset with too short clips from different spatial and temporal aspects is not useful for evaluation. The approximate amount of duration is also specified in the table. Some of the dataset papers do not specify the exact duration and some provide just the number of frames; hence that entry is vacant in the table. 
%Some datasets specially specify the number of normal/abnormal clips. Sometimes a clip has all normal or abnormal frames, and sometimes a clip has both types of frames. 

\begin{figure*}[!htbp]
  \centering
    \includegraphics[scale=0.45]{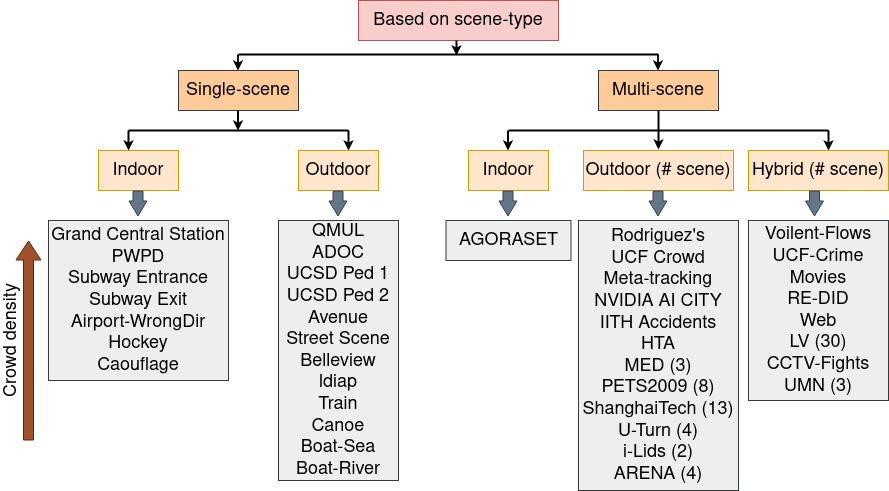}
   \caption{Categorization of video anomaly datasets based on scene-type}
  \label{fig:video_taxonomy_sceneType_based}
\end{figure*}

\begin{figure*}[!htbp]
  \centering
    \includegraphics[scale=0.44]{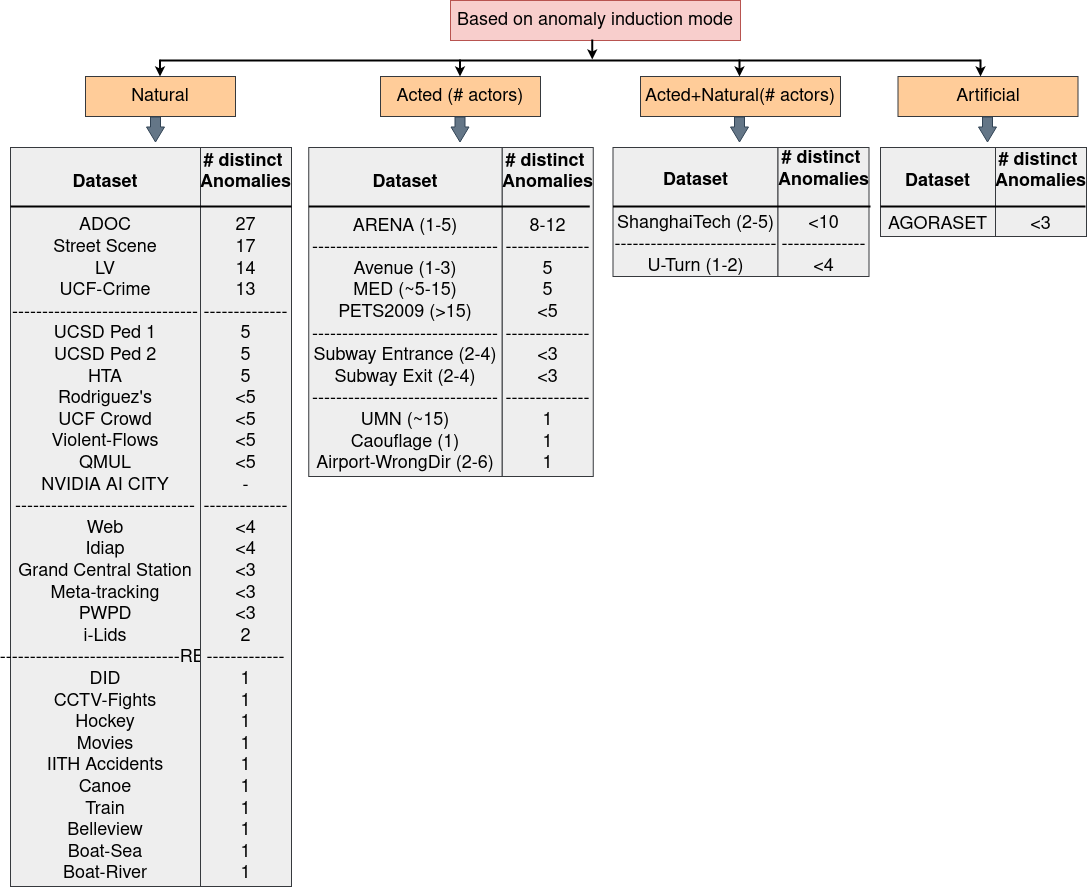}
    \caption{Categorization of video anomaly datasets based on anomaly induction mode}
  \label{fig:video_taxonomy_nature_based}
\end{figure*}

Generally, authors have used clips comprising of only normal frames in training and those with normal and abnormal frames in the testing phase. If the exact number of normal (N), abnormal (A), normal-abnormal (AN) clips are specified by the authors, then we have added this information in the table. Otherwise, the total number of clips is specified there. Some datasets, mainly scraped from online sources (youTube, Pond5, Getty Images), have multiple resolutions; therefore, they need to be scaled on the same scale before processing. Also, for the same reason, these datasets may have multiple FPS. Some datasets have camera motion present too. For some of the datasets, motion is intentionally added to make the dataset more challenging and realistic (surveillance in moving trains such as Train~\cite{zaharescu2010anomalous}, etc., surveillance with head-mounted cameras such as HTA~\cite{singh2020anomalous}, etc.). 
%%%%%%%%%%%%%%%%%%%%%%%%%%%%%%%%%%%%%%%%%%%%%%%%%%%%%%%%%%%%%%%%%%%%%%%%%%%%%
%%%%%%%%%%%%%%%%%%%%%%%%%%%%%%%%%%%%%%%%%%%%%%%%%%%%%%%%%%%%%%%%%%%%%%%%%%%%%%
%\subsection{Dataset classification}
In the following subsections, we present a systematic comparison and remark on publicly available video datasets based on (1) area of application, (2) scene type, (3) anomaly induction mode, and (4) labelling procedure. We describe and compare video datasets through Figures~\ref{fig:video_taxonomy_application_based} to~\ref{fig:video_taxonomy_labeling_based}, following the categorization basis mentioned in Fig.~\ref{fig:taxonomy}. At the end of this section, we provide a more detailed overview of each dataset in Section~\ref{Dataset Overview} with Tables{~\ref{tab:videoTableMain1}} to {\ref{tab:videoTableMain3}}.   
%%%%%%%%%%%%%%%%%%%%%%%%%%%%%%%%%%%%%%%%%%%%%%%%%%%%%%%%%%%%%%%%%%
%%%%%%%%%%%%%%%%%%%%%%%%%%%%%%%%%%%%%%%%%%%%%%%%%%%%%%
\subsection{Area of application}
\begin{figure*}[!htbp]
  \centering
    \includegraphics[scale=0.45]{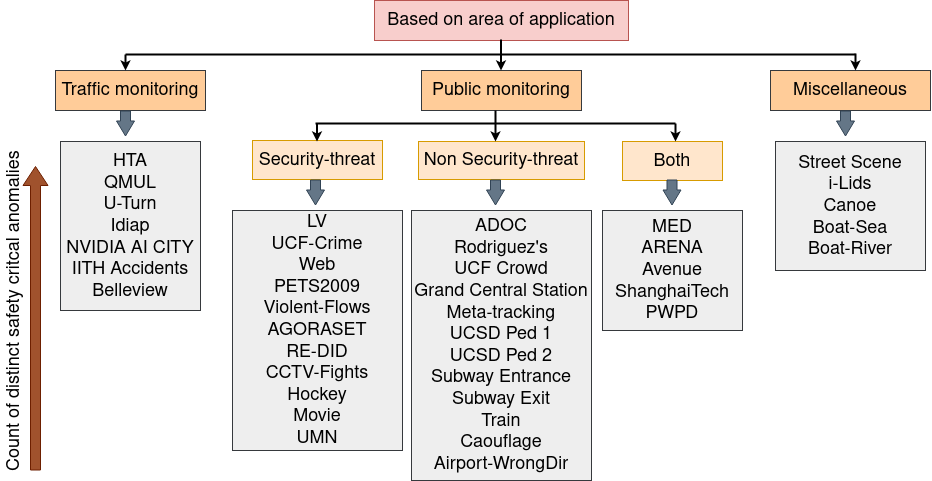}
   \caption{Categorization of video anomaly datasets based on area of application}
  \label{fig:video_taxonomy_application_based}
\end{figure*}

Based on the area of application, the video datasets can be broadly categorized into traffic monitoring and public monitoring applications. Some datasets have a mix of utility for both traffic as well as public monitoring; we keep them in the miscellaneous application category. This categorization is shown in Fig.~\ref{fig:video_taxonomy_application_based}. The datasets built for traffic monitoring application have only traffic-related anomalies and hence owns exclusive application only to traffic surveillance. 
This category has seven datasets, and all of them possess mainly safety-critical anomalies, e.g., accident, tire skidding, wrong turn, etc. On the other hand, the public monitoring application has diversity in the dataset recording location and can be used across indoor to outdoor public place monitoring, e.g., mall, railway station, park, subway, street, etc. Most of the available video anomaly datasets fall into this category. Some of these datasets only possess safety-critical anomalies such as blast, fight, robbery, gunshot, etc., while others possess non-safety-critical anomalies such as walking with a cart in a pedestrian area, throwing papers in the air, walking on grass, etc. Further, some datasets have both types of anomalies. Thus, the public monitoring category is further categorized into `security-threat', `non-security-threat', and `both' sub-categories, based on the criticalness of present anomalies. Some datasets can be used for both traffic as well as public monitoring as they possess anomalies for both categories. We list such datasets under the miscellaneous category. The datasets in each category are listed in order with the count of distinct safety-critical anomalies present in them. A dataset that offers a larger number of distinct safety-critical anomalies is mentioned first.
%%%%%%%%%%%%%%%%%%%%%%%%%%%%%%%%%%%%%%%%%%%%%%%%%%%%%%%%%%%%%%%%%%
%-------------------------------
%-------------------------------
%-------------------------------
%-------------------------------
%-------------------------------
%-------------------------------
%%%%%%%%%%%%%%%%%%%%%%%%%%%%%%%%%%%%%%%%%%%%%%%%%%%%%%%%
\subsection{Scene-type}
Some of the video anomaly datasets are recorded by placing the recording device at one place, while others collect video data from multiple places. Therefore, based on the dataset recording scene, there can be two categories of datasets: single-scene and multi-scene. They can be further categorized based on whether the scene is indoor or outdoor. A dataset recorded indoor or outdoor may help to evaluate the robustness or claim that the framework works in indoor and/or outdoor scenarios. We show the taxonomy for scene-based video datasets categorization via Fig~\ref{fig:video_taxonomy_sceneType_based}. For datasets falling in the multi-scene category, we mention the exact number of locations in parenthesis along with the dataset if provided by the authors of the datasets. Also, the datasets in each sub-category are listed in order of crowd density present in the dataset. Some frameworks are specifically designed to handle crowded scenes, whereas others work only for low density of people; thus, the information of density in a dataset may help to choose suitable datasets for evaluations. 

In the multi-scene category, the datasets mainly possess non-contextual anomalies, which means if an event is anomalous in one scene, it will be anomalous in other scenes as well. For example, fighting with hands is an anomaly for a boxing game scene as well as in a park scene. Therefore, datasets falling in the multi-scene category largely ignore the context. On the other hand, in the case of single-scene datasets, events are regarded as anomalies based on their frequency. An event that is rare for the specific scene is regarded anomaly for that scene only. If the aim is to develop a generic scene monitoring, then single-scene datasets are useful, whereas, for detecting specific anomaly events, multi-scene datasets are suitable.
%%%%%%%%%%%%%%%%%%%%%%%%%%%%%%%%%%%%%%%%%%%%%%%%%%%%%%%%%%%%%%%%%%
\subsection{Anomaly induction mode}
%%Anomaly is a rare phenomena, hence getting variety of natural anomaly samples is challenging. 
Based on the induction mode of anomalies, the datasets can be categorized into four groups, viz., acted, natural, acted as well as natural, and artificial, as shown in Fig.~\ref{fig:video_taxonomy_nature_based}. Since anomaly is a rare phenomenon, it will require a longer waiting time to get the anomalous samples/ events; consequently, researchers have also tried inducing anomalies manually with the help of actors or artificially with the help of computer software. However, the acted events add unrealistic flavor, and hence less complex anomaly events are generated. In Fig.~\ref{fig:video_taxonomy_nature_based}, we also mention the distinct number of anomalies present in each dataset. Further, if the dataset has acted anomalies too, the number of actors involved in inducing anomalies are also reported in parenthesis. 

Generally, for traffic surveillance-related datasets such as HTA~\cite{singh2020anomalous}, QMUL~\cite{loy2008local}, Idiap~\cite{varadarajan2009topic}, IITH Accidents~\cite{singh2018deep}, etc., getting natural anomaly samples is easier as the anomalies in these datasets, which include accidents, illegal turns, or other traffic rule violations, occur a bit frequently. Also, safety-critical events in public monitoring such as fights, blasts, robbery, gunshots are easier to collect from CCTV footage, movies, or YouTube. Hence, datasets such as UCF-Crime~\cite{sultani2018real}, LV~\cite{leyva2017lv}, CCTV-Fights~\cite{perez2019detection}, Movie~\cite{nievas2011violence}, etc., have natural anomalies. On the other hand, datasets having fewer safety-critical anomalies are built by inducing acted events too. Some of these include ARENA~\cite{pets2014}, Avenue~\cite{lu2013abnormal}, MED~\cite{rabiee2016novel}, ShanghaiTech~\cite{luo2017revisit}, etc.
%====================================
%%%%Getting anomalous natural events for human-centric anomaly datasets is difficult; therefore, many of the datasets have acted anomalies. The acted events add unrealistic flavor, and hence less complex anomaly events are generated. Generally, traffic surveillance-related datasets get natural anomaly samples as accidents, illegal turns, or other traffic rule violations are a bit common. Anomalies in the densely crowded scene are also easier to spot. However, very few types of anomalies are found in specific anomaly categories. On the contrary, heterogeneous datasets for public surveillance offer more diverse types of anomalies. We also observe that the amount of anomalies in a dataset has increased in recent years (such as LV, UCF-Crime, Street Scene, ADOC).
%-------------------------------
\begin{figure*}[!htbp]
  \centering
    \includegraphics[scale=0.5]{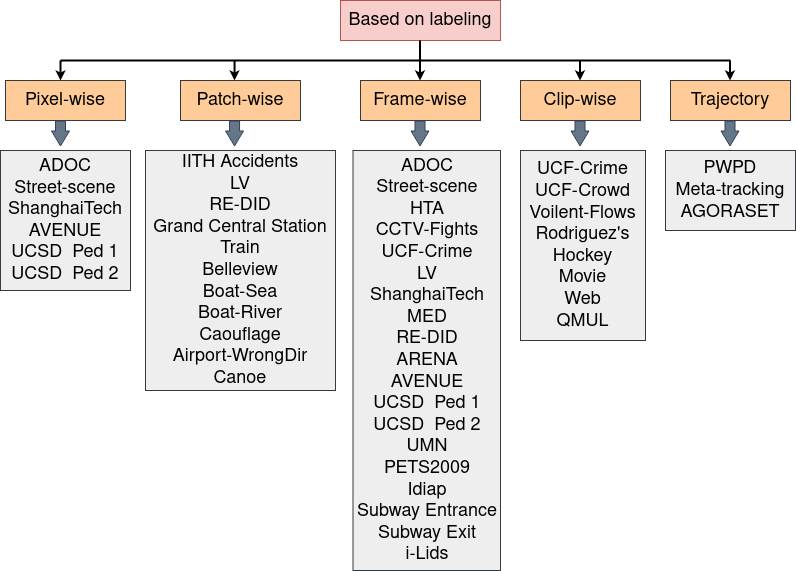}
   \caption{Categorization of video anomaly datasets based on labeling procedure}
  \label{fig:video_taxonomy_labeling_based}
\end{figure*}
%-------------------------------
\subsection{Labelling procedure}
The existing video anomaly datasets have diversity in terms of the type of labeling too. We categorize them based on the available annotation type, as shown in Fig.~\ref{fig:video_taxonomy_labeling_based}. Some datasets give only clip-level annotation; if a clip is annotated anomalous, then it is assumed that anomaly is present or spanned to all the frames in the clip. The assumptions hold true in most cases, especially for small databases, but it is not scalable to validate this assumption for larger datasets. Hence, some frames may be wrongly annotated. Frame-level annotations are more precise and accurate~\cite{ramachandra2020street}. For some datasets, pixel-level annotations are also provided, which are helpful for models which aim to precisely localize the occurrence of an anomaly in the spatial domain along with the temporal domain. Pixel-wise annotations are too precise and hence difficult to do manually; therefore, some datasets such as Train~\cite{zaharescu2010anomalous}, Belleview~\cite{zaharescu2010anomalous}, Boat-Sea~\cite{zaharescu2010anomalous}, BoatRiver~\cite{zaharescu2010anomalous}, Airport-WrongDir~\cite{zaharescu2010anomalous}, etc. first divide the frame into grids (a grid can have multiple pixels) and then annotate the grids that can be served for spatial anomaly localization. Some also provide the rectangle bounding boxes around the area of interest, i.e., anomalous object/person. Some datasets, especially for people tracking crowd datasets, e.g., PWPD~\cite{yi2015understanding}, meta-tracking~\cite{jodoin2013meta}, etc., provide the trajectories and anomaly information on these.
%%%%%%%%%%%%%%%%%%%%%%%%%%%%%%%%%%%%%%%%%%%%%%%%%%%%%%%%%%%%%%%%%%
\subsection{Dataset Overview}\label{Dataset Overview}
A brief introduction of all the publicly available video anomaly datasets is given through Tables~\ref{tab:videoTableMain1} to \ref{tab:videoTableMain3}. The datasets are listed in chronological order of their release year. Some examples of anomalies present and the recording location of the dataset are also mentioned. Information about dataset collection scene, i.e., whether it is a mall or an airport or a traffic junction, etc., may help to choose scene-specific dataset for the evaluation of scene-specific surveillance frameworks. The table also mentions the area of application which gives an idea of different applications where the dataset has been used for bench-marking. Visuals from the dataset, i.e., a sample normal and an abnormal frame from the datasets, are also shown.
%%%%%%%%%%%%%%%%%%%%%%%%%%%%%%%%%%%%%%%%%%%%%%%%%%%%%%%%%%%%%%%%%%
%=====================================================
\begin{table*}
  \caption{Video anomaly datasets: primary information}
  \label{tab:videoTableMain1}
  \centering
  \resizebox{.99\textwidth}{!}{
  \begin{tabular}{|c|c|c|c|c|}
    \hline
     Dataset&\begin{tabular}[c]{@{}c@{}}Anomalies: \\collection scenario \end{tabular} & Applications&Normal image example&Anomaly image example\\\hline 
    
     \begin{tabular}[c]{@{}c@{}} i-Lids~\cite{iLid}\\(2007)\end{tabular} &
     \begin{tabular}[c]{@{}c@{}}abandon bag,\\illegally parked vehicle: \\at railway station,\\road\end{tabular}&
     \begin{tabular}[c]{@{}c@{}} abandon baggage \\detection~\cite{li2009robust},\\ traffic surveillance~\cite{cui2011abnormal}\end{tabular} &  
     \begin{minipage}{.2\textwidth}
      \includegraphics[scale=0.12]{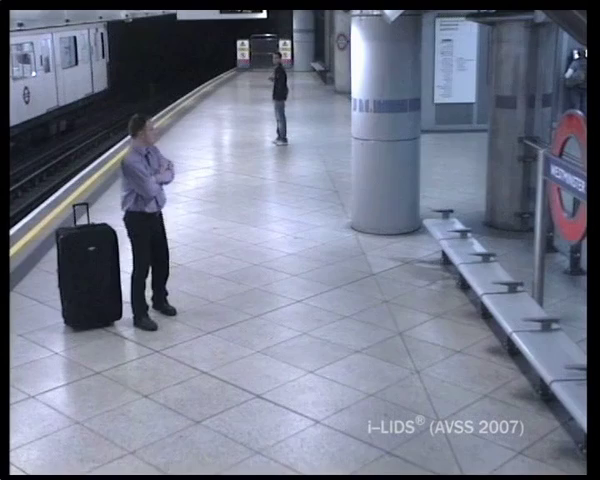}
    \end{minipage}&
    \begin{minipage}{.2\textwidth}
      \includegraphics[scale=0.12]{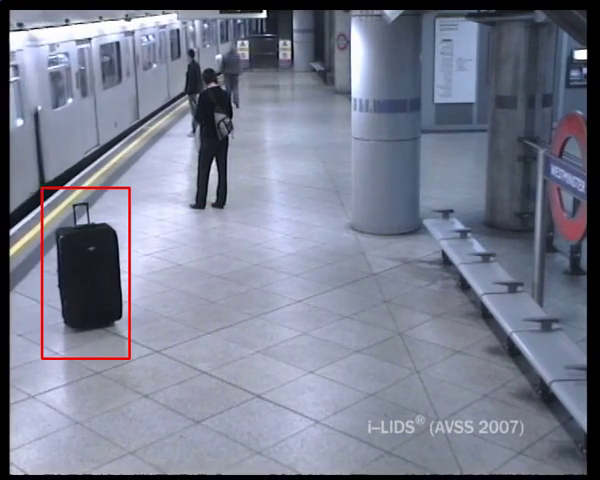}
    \end{minipage}\\ \hline

     \begin{tabular}[c]{@{}c@{}} QMUL~\cite{loy2008local}\\(2008)\end{tabular}&
     \begin{tabular}[c]{@{}c@{}}unusual traffic trajectory,\\ rare behaviour of \\vehicles: at road \\(traffic junction)\end{tabular}&
       \begin{tabular}[c]{@{}c@{}} vehicle tracking~\cite{santhosh2019trajectory}, \\trajectory classification~\cite{santhosh2021vehicular},\\anomaly detection~\cite{loy2011detecting,varadarajan2017active,kaltsa2018multiple}\end{tabular} &
     \begin{minipage}{.2\textwidth}
      \includegraphics[scale=0.2]{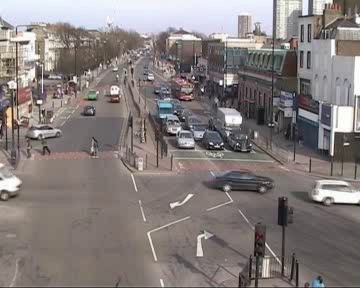}
    \end{minipage}&
    \begin{minipage}{.2\textwidth}
      \includegraphics[scale=0.2]{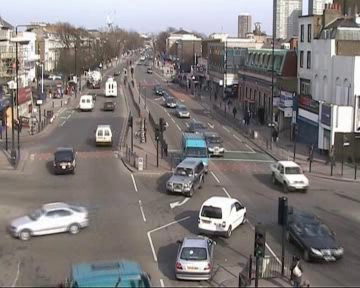}
    \end{minipage}\\ \hline
     
    \begin{tabular}[c]{@{}c@{}} Canoe~\cite{jodoin2008modeling}\\(2008)\end{tabular}  &
    canoe: in river&
    anomaly detection~\cite{zaharescu2010anomalous,dos2019generalization}& 
     \begin{minipage}{.2\textwidth}
      \includegraphics[scale=0.24]{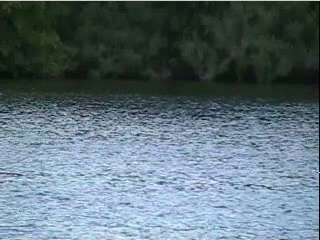}
    \end{minipage}&
    \begin{minipage}{.2\textwidth}
      \includegraphics[scale=0.24]{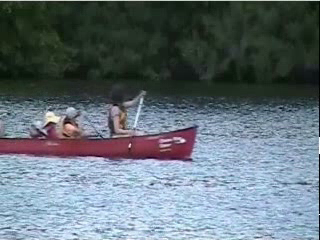}
    \end{minipage}\\ \hline
    
    \begin{tabular}[c]{@{}c@{}} Subway Entrance\\~\cite{adam2008robust}(2008)\end{tabular}&
    \begin{tabular}[c]{@{}c@{}} wrong direction:\\ at subway entrance\end{tabular}&
    \begin{tabular}[c]{@{}c@{}}Anomaly detection~\cite{cong2011sparse,kratz2009anomaly}, \\abnormal behavior\\ modeling~\cite{popoola2012video,javan2013online}\end{tabular} & 
      \begin{minipage}{.2\textwidth}
      \includegraphics[scale=0.15]{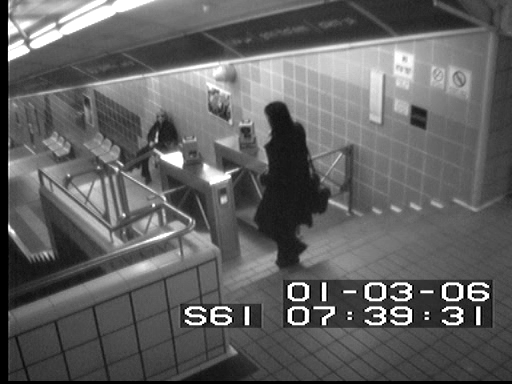}
    \end{minipage}&
    \begin{minipage}{.2\textwidth}
      \includegraphics[scale=0.15]{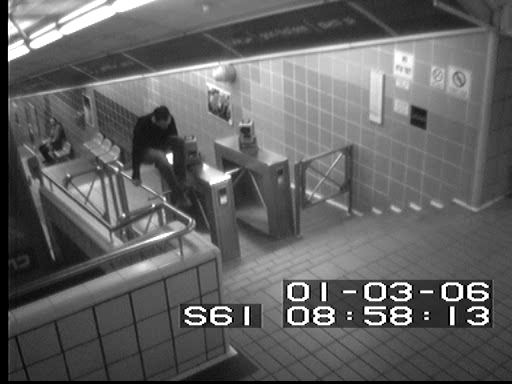}
    \end{minipage}\\ \hline
     
    \begin{tabular}[c]{@{}c@{}} Subway Exit\\~\cite{adam2008robust}(2008)\end{tabular}&
    \begin{tabular}[c]{@{}c@{}} wrong direction:\\ at subway exit\end{tabular}&
    \begin{tabular}[c]{@{}c@{}}Anomaly detection~\cite{cong2011sparse,kratz2009anomaly}, \\abnormal behavior \\modeling~\cite{popoola2012video,javan2013online}\end{tabular} & 
     \begin{minipage}{.2\textwidth}
      \includegraphics[scale=0.43]{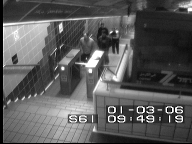}
    \end{minipage}&
    \begin{minipage}{.2\textwidth}
      \includegraphics[scale=0.43]{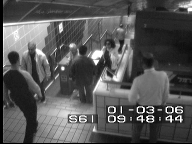}
    \end{minipage}\\ \hline
    
    \begin{tabular}[c]{@{}c@{}}UMN~\cite{xyz} \\(2009)\end{tabular}  &
    \begin{tabular}[c]{@{}c@{}}run: in field, \\courtyard, hallway\end{tabular}  &
    \begin{tabular}[c]{@{}c@{}}anomaly detection~\cite{mahadevan2010anomaly,cong2011sparse}, \\abnormal crowd behaviour~\cite{mousavi2015analyzing,su2013large}, \\crowd aggregation detection~\cite{xu2017efficient},\\ crowd escape detection~\cite{wu2013bayesian}\end{tabular} & 
     \begin{minipage}{.2\textwidth}
      \includegraphics[scale=0.27]{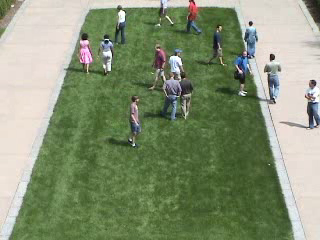}
    \end{minipage}&
    \begin{minipage}{.2\textwidth}
      \includegraphics[scale=0.27]{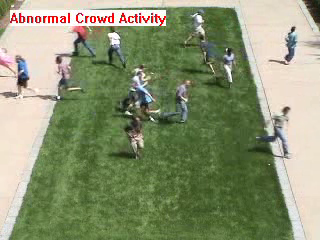}
    \end{minipage}\\ \hline
    
     \begin{tabular}[c]{@{}c@{}}Web~\cite{xyz}\\(2009)\end{tabular}    &
     \begin{tabular}[c]{@{}c@{}} panic-escape, \\clashing/fighting: \\at multiple \\location \end{tabular}&
     anomaly detection~\cite{zhu2014sparse}&  
      \begin{minipage}{.2\textwidth}
      \includegraphics[scale=0.26]{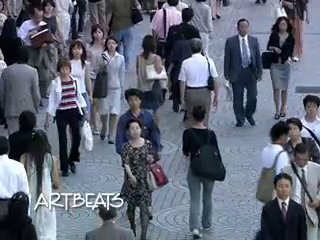}
    \end{minipage}&
    \begin{minipage}{.2\textwidth}
      \includegraphics[scale=0.16]{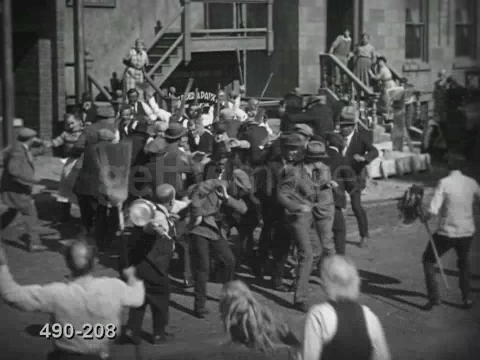}
    \end{minipage}\\ \hline
     
    \begin{tabular}[c]{@{}c@{}}PETS2009~\cite{ferryman2009overview}\\(2009)\end{tabular}  & 
    \begin{tabular}[c]{@{}c@{}}run, \\panic: in university \end{tabular}&
    \begin{tabular}[c]{@{}c@{}}tracking~\cite{leal2011everybody,chu2013tracking},\\ crowd profiling/counting~\cite{loy2013crowd,fradi2015towards},\\ crowd analysis~\cite{su2013large}, \\human detection~\cite{bolme2009simple,conde2013hogg}, \\person re-identification~\cite{yang2011person}, \\crowd escape behaviour detection~\cite{wu2013bayesian}\end{tabular} & 
     \begin{minipage}{.2\textwidth}
      \includegraphics[scale=0.11]{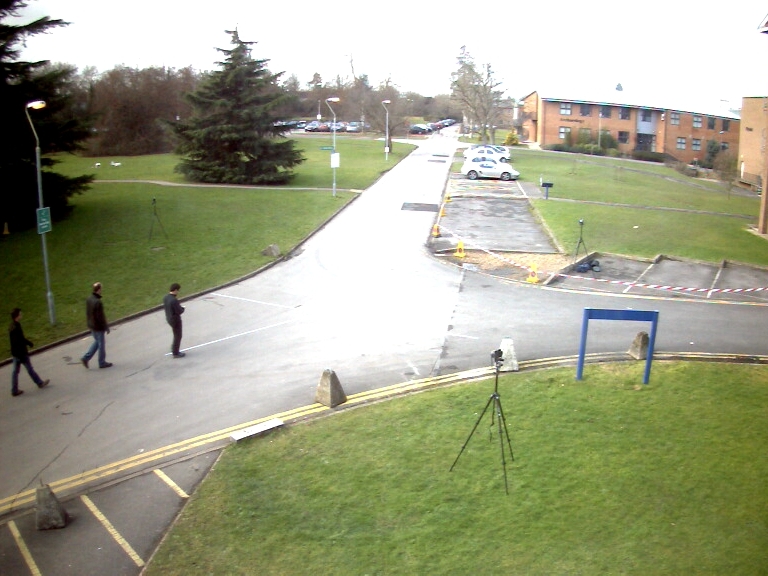}
    \end{minipage}&
    \begin{minipage}{.2\textwidth}
      \includegraphics[scale=0.11]{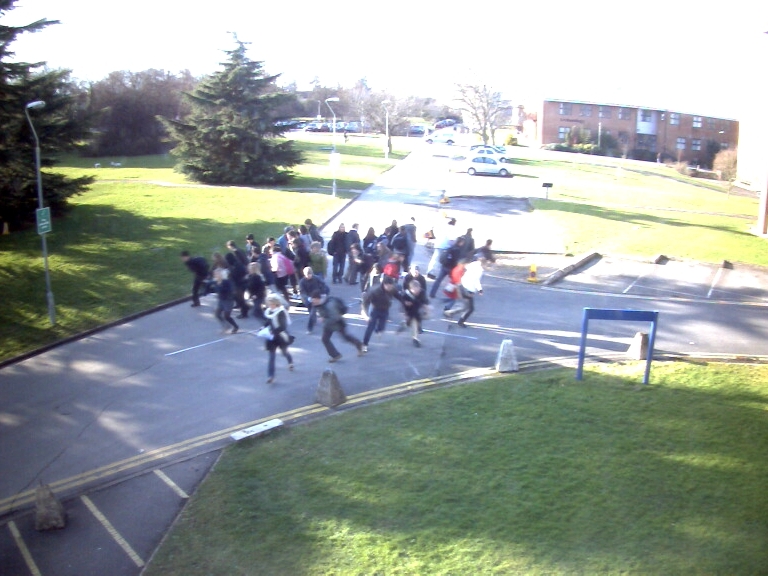}
    \end{minipage}\\ \hline
    
    \begin{tabular}[c]{@{}c@{}}U-Turn~\cite{benezeth2009abnormal}\\(2009)\end{tabular} &
    \begin{tabular}[c]{@{}c@{}}illegal u-turns, \\running, abandon\\ baggage: at road \\(intersection), university\end{tabular}&
    \begin{tabular}[c]{@{}c@{}}traffic based \\anomaly detection~\cite{zhou2016spatial,benezeth2011abnormality,li2013anomaly}\end{tabular} &  
     \begin{minipage}{.2\textwidth}
      \includegraphics[scale=0.2]{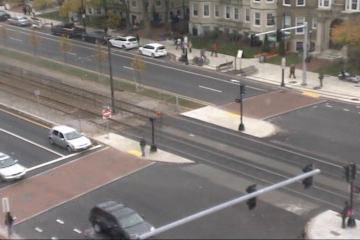}
    \end{minipage}&
    \begin{minipage}{.2\textwidth}
      \includegraphics[scale=0.2]{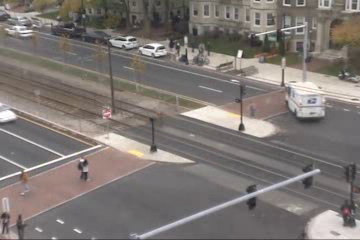}
    \end{minipage}\\ \hline
     
     \begin{tabular}[c]{@{}c@{}}Idiap~\cite{varadarajan2009topic}\\(2009)\end{tabular}  &
     \begin{tabular}[c]{@{}c@{}}wrong road crossing, \\wrong vehicle parking,\\ etc.,: at road\end{tabular}&
      \begin{tabular}[c]{@{}c@{}}anomaly detection~\cite{varadarajan2017active,kaltsa2018multiple},\\ recurrent activity mining~\cite{varadarajan2013sequential}\end{tabular} &  
     \begin{minipage}{.2\textwidth}
      \includegraphics[scale=0.12]{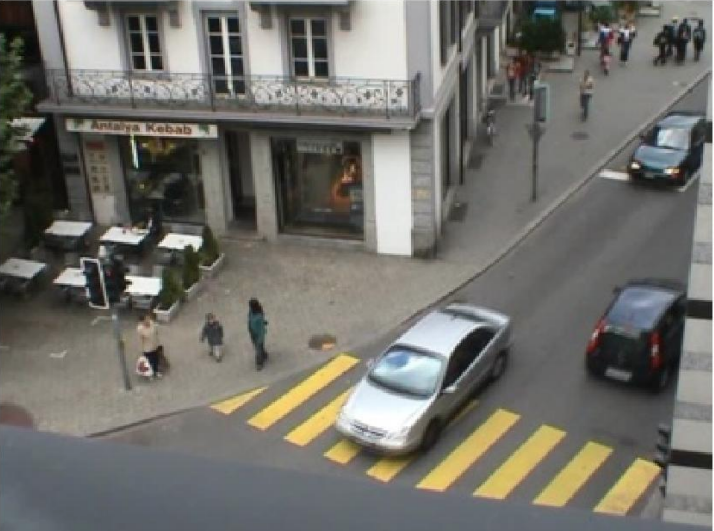}
    \end{minipage}&
    \begin{minipage}{.2\textwidth}
      \includegraphics[scale=0.12]{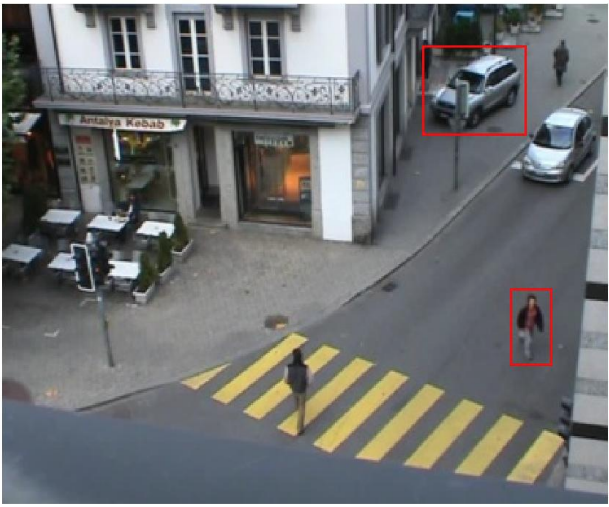}
    \end{minipage}\\ \hline
    
     \begin{tabular}[c]{@{}c@{}}USCD Ped1~\cite{mahadevan2010anomaly}\\(2010) \end{tabular} & 
   \begin{tabular}[c]{@{}c@{}} bikers, small carts,\\ walking across walkways:\\ at campus (walkway\\ at UCSD)\end{tabular}  &
   \begin{tabular}[c]{@{}c@{}}crowd profiling/counting~\cite{loy2013crowd,xu2016crowd},\\anomaly detection~\cite{ravanbakhsh2018plug,cong2013abnormal},\\ crowd density estimation~\cite{pham2015count} \end{tabular} &   
    \begin{minipage}{.2\textwidth}
      \includegraphics[scale=0.35]{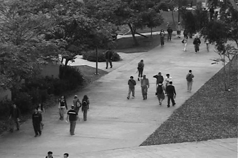}
    \end{minipage}&
    \begin{minipage}{.2\textwidth}
      \includegraphics[scale=0.35]{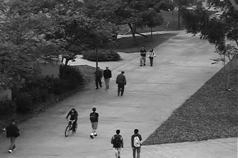}
    \end{minipage}\\ \hline
    
    \begin{tabular}[c]{@{}c@{}}USCD Ped2~\cite{mahadevan2010anomaly}\\(2010) \end{tabular} &
    \begin{tabular}[c]{@{}c@{}} bikers, small carts,\\ walking across walkways:\\ at campus (walkway\\ at UCSD)\end{tabular}&
    \begin{tabular}[c]{@{}c@{}}anomaly detection~\cite{sabokrou2018deep,xu2017detecting}, \\crowd profiling/counting~\cite{loy2013crowd},\\action classification~\cite{zhong2019graph} \end{tabular} & 
    \begin{minipage}{.2\textwidth}
      \includegraphics[scale=0.23]{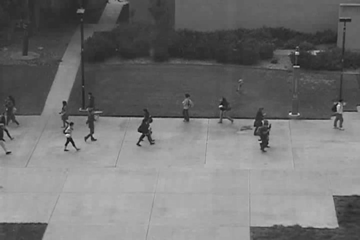}
    \end{minipage}&
    \begin{minipage}{.2\textwidth}
      \includegraphics[scale=0.23]{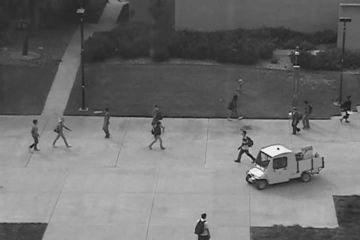}
    \end{minipage}\\ \hline
\end{tabular}}
\end{table*}
  %%%%%%%%%%%%%%%%%%%%%%%%%%%%%%%%%%%%%%%%%%%%%%%   
\begin{table*}
  \caption{Continued from Table \ref{tab:videoTableMain1}: Video anomaly datasets: primary information}
  \label{tab:videoTableMain2}
  \centering
  \resizebox{.99\textwidth}{!}{
  \begin{tabular}{|c|c|c|c|c|}
    \hline
     Dataset& \begin{tabular}[c]{@{}c@{}}Anomalies: \\collection scenario \end{tabular} & Applications&Normal image example&Anomaly image example\\\hline

    \begin{tabular}[c]{@{}c@{}}Train~\cite{zaharescu2010anomalous}\\(2010)\end{tabular}  &
    \begin{tabular}[c]{@{}c@{}}people movement:\\ inside train\end{tabular}  &
    anomaly detection~\cite{xiao2015learning,xu2017detecting,dos2019generalization}&
      \begin{minipage}{.2\textwidth}
      \includegraphics[scale=0.13]{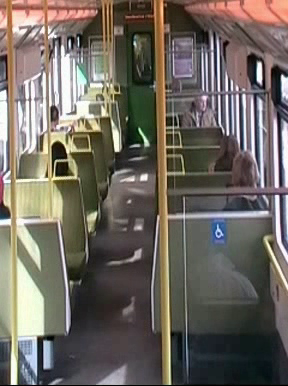}
    \end{minipage}&
    \begin{minipage}{.2\textwidth}
      \includegraphics[scale=0.13]{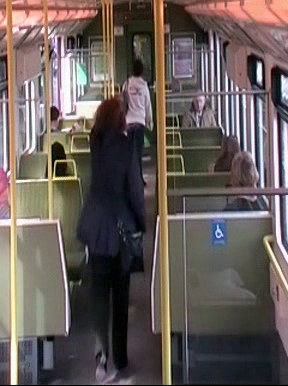}
    \end{minipage}\\ \hline
     
    \begin{tabular}[c]{@{}c@{}}Belleview~\cite{zaharescu2010anomalous}\\(2010)\end{tabular} &
     \begin{tabular}[c]{@{}c@{}}illegal turns: at \\road (intersection)\end{tabular}&
    anomaly detection~\cite{xiao2015learning,dos2019generalization}&
     \begin{minipage}{.2\textwidth}
      \includegraphics[scale=0.2]{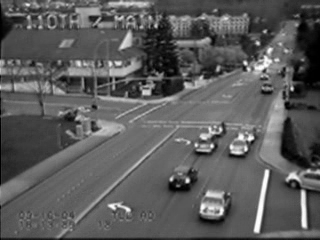}
    \end{minipage}&
    \begin{minipage}{.2\textwidth}
      \includegraphics[scale=0.2]{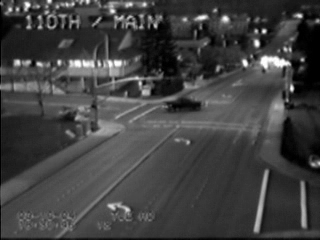}
    \end{minipage}\\ \hline
    
    \begin{tabular}[c]{@{}c@{}}Boat-Sea~\cite{zaharescu2010anomalous}\\(2010)\end{tabular}  &
    boat: in sea&
    anomaly detection~\cite{roshtkhari2013line,dos2019generalization}&
     \begin{minipage}{.2\textwidth}
      \includegraphics[scale=0.08]{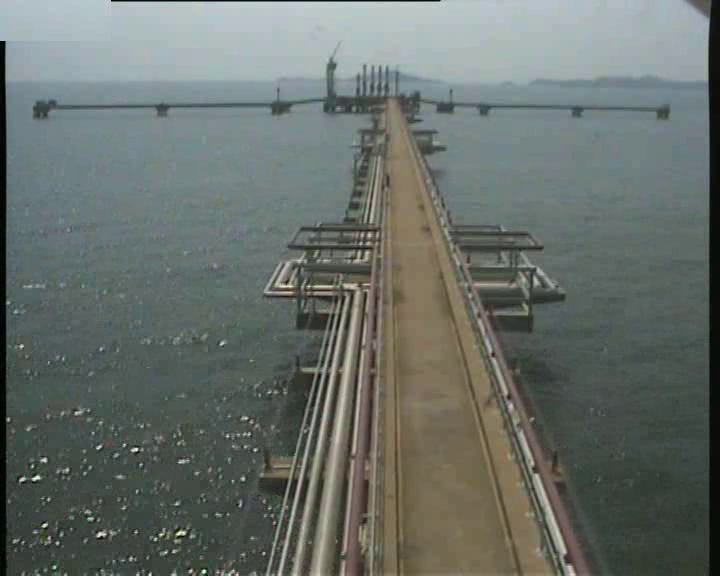}
    \end{minipage}&
    \begin{minipage}{.2\textwidth}
      \includegraphics[scale=0.08]{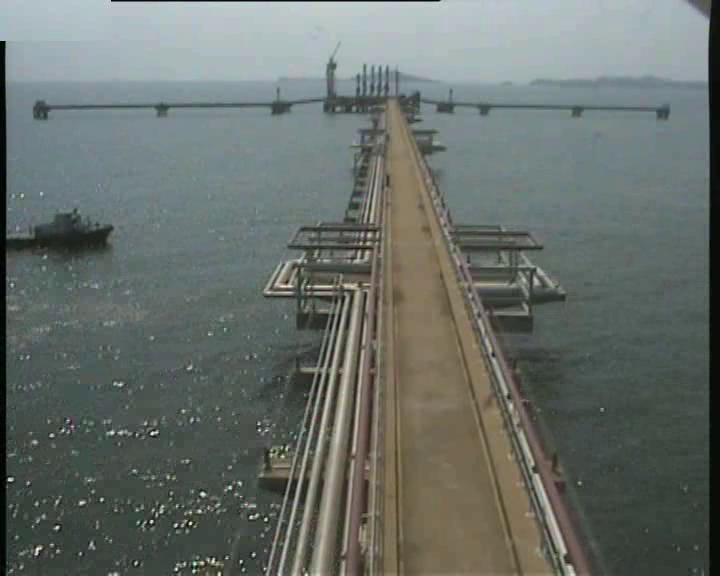}
    \end{minipage}\\ \hline
    
    \begin{tabular}[c]{@{}c@{}}Boat-River~\cite{zaharescu2010anomalous}\\(2010)\end{tabular} &
    boat: in river&
    anomaly detection~\cite{dos2019generalization,javan2013online}&
     \begin{minipage}{.2\textwidth}
      \includegraphics[scale=0.08]{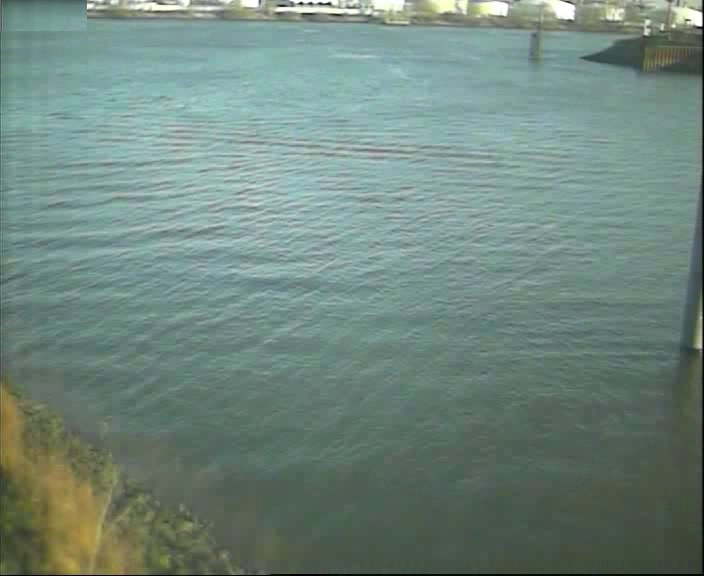}
    \end{minipage}&
    \begin{minipage}{.2\textwidth}
      \includegraphics[scale=0.08]{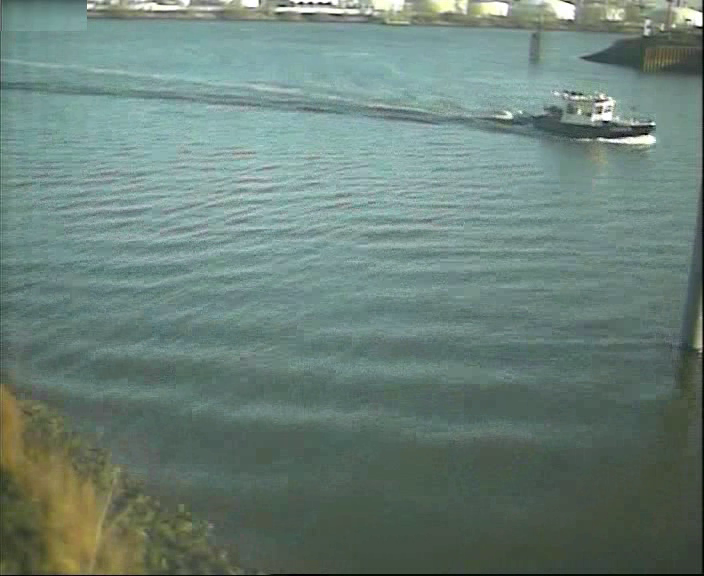}
    \end{minipage}\\ \hline
    
    \begin{tabular}[c]{@{}c@{}}Caouflage~\cite{zaharescu2010anomalous}\\(2010)\end{tabular}  &
    \begin{tabular}[c]{@{}c@{}}person in Caouflage:\\ in room\end{tabular} &
    anomaly detection~\cite{javan2013online}&
     \begin{minipage}{.2\textwidth}
      \includegraphics[scale=0.2]{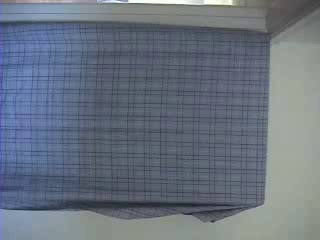}
    \end{minipage}&
    \begin{minipage}{.2\textwidth}
      \includegraphics[scale=0.2]{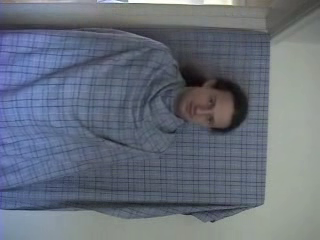}
    \end{minipage}\\ \hline
    
    \begin{tabular}[c]{@{}c@{}}Airport-WrongDir~\cite{zaharescu2010anomalous}\\(2010)\end{tabular}  &
    \begin{tabular}[c]{@{}c@{}}movement in wrong \\direction: at security\\ check-point\end{tabular}&
     anomaly detection~\cite{joshi2021cnn,javan2013online}&
     \begin{minipage}{.2\textwidth}
      \includegraphics[scale=0.2]{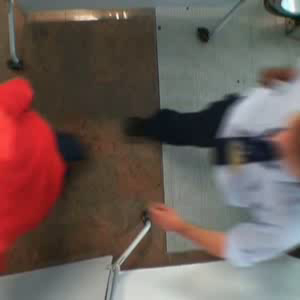}
    \end{minipage}&
    \begin{minipage}{.2\textwidth}
      \includegraphics[scale=0.2]{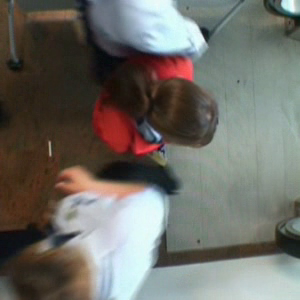}
    \end{minipage}\\ \hline

    \begin{tabular}[c]{@{}c@{}}Rodriguez's~\cite{rodriguez2011data}\\(2011)\end{tabular}   
    & \begin{tabular}[c]{@{}c@{}}anomalous trajectory: \\from multiple \\scenarios\end{tabular} 
    &\begin{tabular}[c]{@{}c@{}}crowd profiling/counting~\cite{fradi2015towards},\\crowd saliency detection~\cite{lim2014crowd},\\ tracking~\cite{chu2013tracking},\\
crowd analysis~\cite{bera2016realtime}, \\crowd segmentation~\cite{kok2016grcs}\end{tabular}   &
     \begin{minipage}{.2\textwidth}
      \includegraphics[scale=0.5]{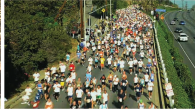}
    \end{minipage}&
    \begin{minipage}{.2\textwidth}
      \includegraphics[scale=0.25]{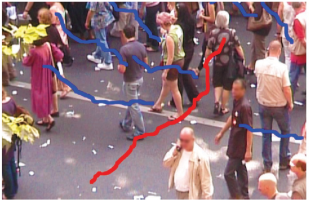}
    \end{minipage}\\ \hline
    
     \begin{tabular}[c]{@{}c@{}}Hockey~\cite{nievas2011violence}\\(2011)\end{tabular}  &
     \begin{tabular}[c]{@{}c@{}} fight: at ice \\hockey game \end{tabular} &
     violence detection~\cite{gao2016violence,xu2018violent,cheng2021rwf}&
    available upon request&
    \begin{minipage}{.2\textwidth}
      \includegraphics[scale=0.2]{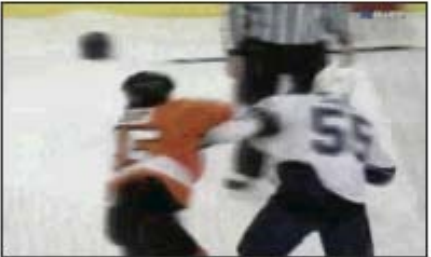}
    \end{minipage}
    \\ \hline
    
    \begin{tabular}[c]{@{}c@{}}Movie~\cite{nievas2011violence}\\(2011)\end{tabular}
    & \begin{tabular}[c]{@{}c@{}}fight: in \\movie clips\end{tabular}
    &violence detection~\cite{xu2018violent,cheng2021rwf,song2019novel,khan2019cover}&
    available upon request&
    \begin{minipage}{.2\textwidth}
      \includegraphics[scale=0.2]{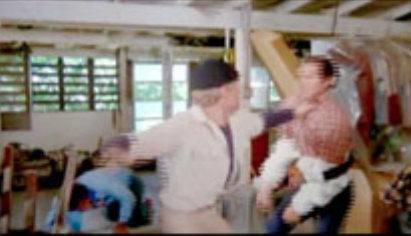}
    \end{minipage}\\ \hline
    
    \begin{tabular}[c]{@{}c@{}}UCF Crowd~\cite{solmaz2012identifying}\\(2012)\end{tabular} &
     \begin{tabular}[c]{@{}c@{}}anomalous trajectory: \\from multiple \\scenarios\end{tabular}&
      \begin{tabular}[c]{@{}c@{}}abnormal crowd detection~\cite{zhang2014social}, \\crowd profiling/counting~\cite{fradi2015towards}, \\crowd saliency detection~\cite{lim2014crowd},\\crowd segmentation~\cite{li2016crowd}\end{tabular} &
     \begin{minipage}{.2\textwidth}
      \includegraphics[scale=0.13]{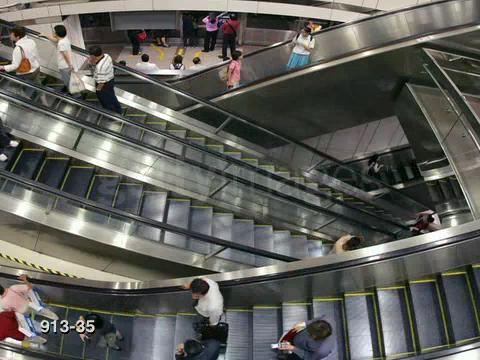}
    \end{minipage}&
    \begin{minipage}{.2\textwidth}
      \includegraphics[scale=0.13]{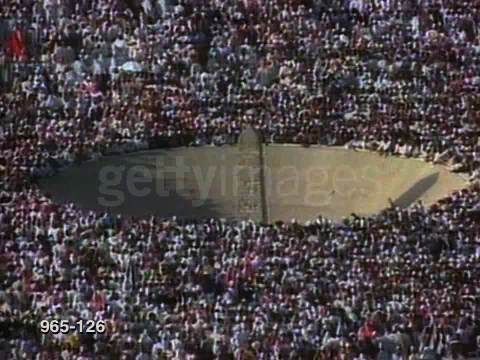}
    \end{minipage}\\ \hline
    
    \begin{tabular}[c]{@{}c@{}}Violent-Flows~\cite{hassner2012violent}\\(2012)\end{tabular}  
    & \begin{tabular}[c]{@{}c@{}}crowd violence: \\at multiple scene \end{tabular}  &
       \begin{tabular}[c]{@{}c@{}}abnormal crowd behaviour\\ detection~\cite{mousavi2015analyzing},\\violence detection~\cite{song2019novel,xu2018violent}\end{tabular} &
     \begin{minipage}{.2\textwidth}
      \includegraphics[scale=0.1]{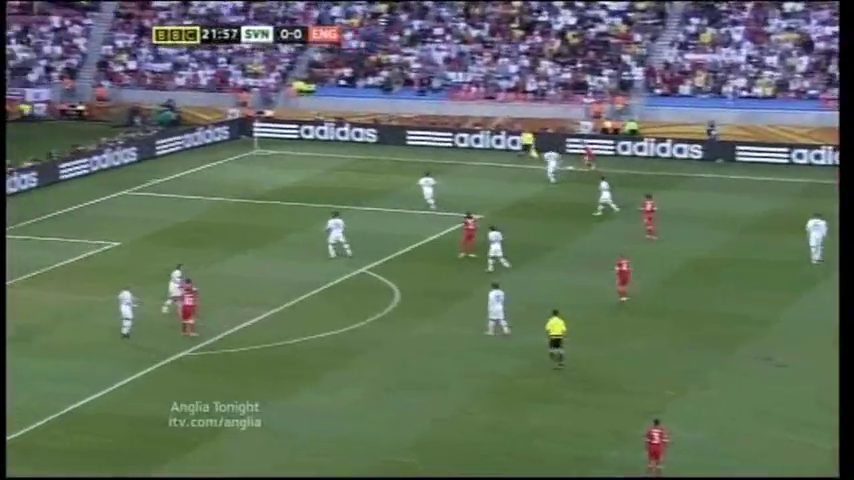}
    \end{minipage}&
    \begin{minipage}{.2\textwidth}
      \includegraphics[scale=0.2]{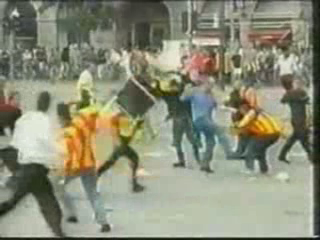}
    \end{minipage}\\ \hline
    
    \begin{tabular}[c]{@{}c@{}}Grand Central \\Station~\cite{zhou2012understanding} \\(2012)\end{tabular} &
    \begin{tabular}[c]{@{}c@{}}rare walking \\pattern: at \\terminal station\end{tabular} &
     \begin{tabular}[c]{@{}c@{}}pedestrian trajectory prediction~\cite{xu2018encoding},\\ tracking~\cite{maksai2017non,santhosh2019trajectory}\\
        pedestrian behavior modeling~\cite{yi2016pedestrian}, \\
        crowd counting~\cite{xu2016crowd,saugun2017novel}\\
        crowd behavior analysis~\cite{zhong2015learning,zhou2015learning},\\
        human re-identification~\cite{assari2016human}\end{tabular}&
      \begin{minipage}{.2\textwidth}
      \includegraphics[scale=0.15]{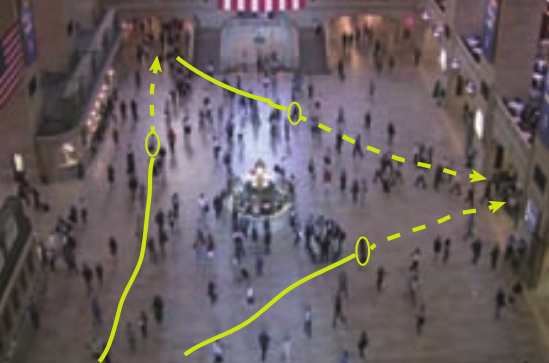}
    \end{minipage}&
    trajectory based\\ \hline
     
    \begin{tabular}[c]{@{}c@{}}AGORASET~\cite{allain2012agoraset}\\(2012)\end{tabular} &
      \begin{tabular}[c]{@{}c@{}}evacuation, \\dispersion: at \\multiple scene\end{tabular} &
       \begin{tabular}[c]{@{}c@{}}panic behaviour detection~\cite{shehab2019statistical},\\
      crowd flow tracking~\cite{fagette2013particle}, \\
        crowd motion classification~\cite{basset2013frame},\\
        crowd behaviour analysis~\cite{pennisi2016online}\end{tabular}&
     \begin{minipage}{.2\textwidth}
      \includegraphics[scale=0.1]{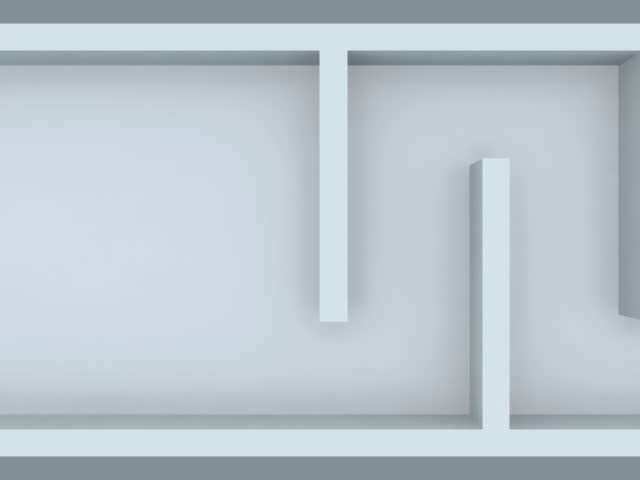}
    \end{minipage}&
    \begin{minipage}{.2\textwidth}
      \includegraphics[scale=0.1]{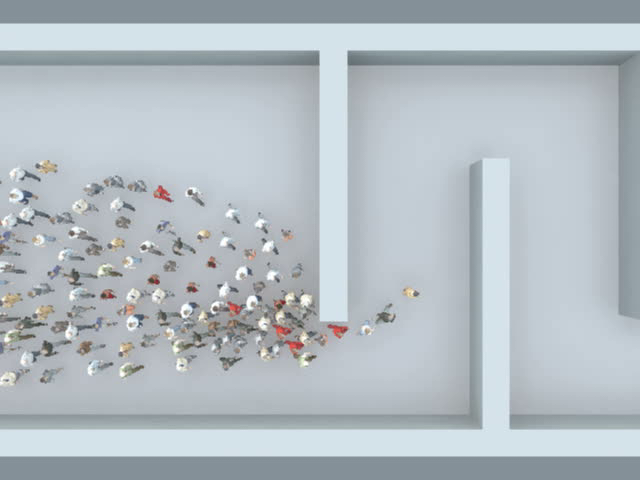}
    \end{minipage}\\ \hline
\end{tabular}}
\end{table*}
%%%%%%%%%%%%%%%%%%%%%%%%%%%%%%%%%%%%%%%%%%%%%%
\begin{table*}
  \caption{Continued from Table \ref{tab:videoTableMain2}: Video anomaly datasets: primary information}
  \label{tab:videoTableMain3}
  \centering
  \resizebox{.99\textwidth}{!}{
  \begin{tabular}{|c|c|c|c|c|}
    \hline
    Dataset&\begin{tabular}[c]{@{}c@{}}Anomalies: \\collection scenario \end{tabular} & Applications&Normal image example&Anomaly image example\\\hline

    \begin{tabular}[c]{@{}c@{}}Meta-tracking\\~\cite{jodoin2013meta} (2013)\end{tabular} &
    \begin{tabular}[c]{@{}c@{}}anomalous trajectory: \\ at multiple location\end{tabular} &
     \begin{tabular}[c]{@{}c@{}}  tracking~\cite{kim2015interactive,li2016measuring},\\
       anomalous walking pattern detection~\cite{kim2015interactive}\\
        measuring collectiveness~\cite{li2016measuring},\\
        crowd segmentation~\cite{fan2018adaptive}\end{tabular} &
     \begin{minipage}{.2\textwidth}
      \includegraphics[scale=0.13]{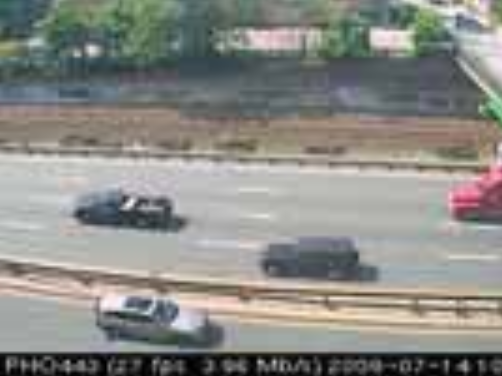}
    \end{minipage}&
    trajectory based\\ \hline
    
    \begin{tabular}[c]{@{}c@{}}Avenue~\cite{lu2013abnormal}\\(2013)\end{tabular}  
    &\begin{tabular}[c]{@{}c@{}}loitering, \\running,\\ throwing objects, \\new object: in \\campus (avenue)\end{tabular}&
     anomaly detection~\cite{gong2019memorizing,hasan2016learning} &  
     \begin{minipage}{.2\textwidth}
      \includegraphics[scale=0.15]{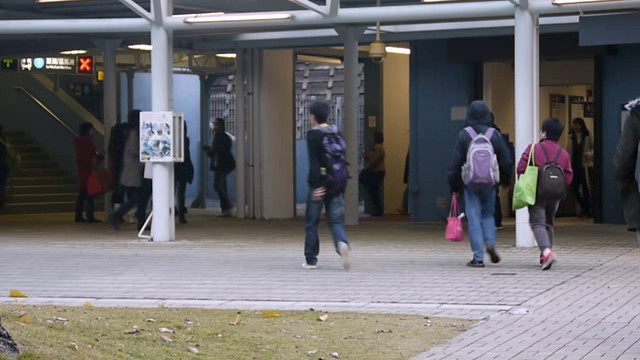}
    \end{minipage}&
    \begin{minipage}{.2\textwidth}
      \includegraphics[scale=0.15]{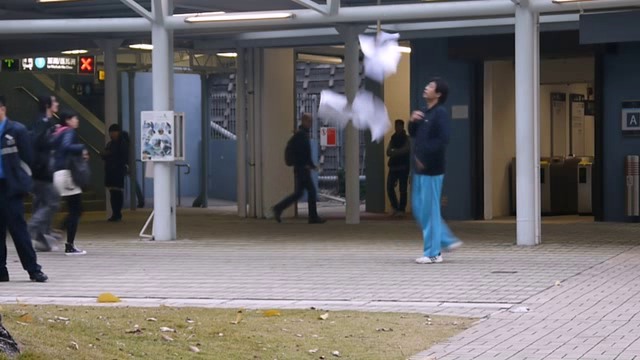}
    \end{minipage}\\ \hline
    
     \begin{tabular}[c]{@{}c@{}}ARENA~\cite{pets2014}\\(2014)\end{tabular}  &
     \begin{tabular}[c]{@{}c@{}}abnormal behaviour, \\threats: in \\campus (university\\of Reading)\end{tabular}&
     \begin{tabular}[c]{@{}c@{}}abnormal activity/behaviour \\detection~\cite{patino2016detecting,burghouts2014complex,patino2014multiresolution},
    group walking event\\ detection~\cite{bastani2015online}, 
    Human Full-Body/\\Body-Parts detection and tracking~\cite{chen2014improved} \end{tabular} &  
    \begin{minipage}{.2\textwidth}
      \includegraphics[scale=0.27]{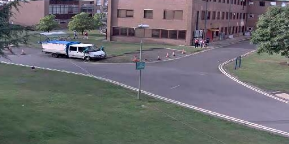}
    \end{minipage}&
    \begin{minipage}{.2\textwidth}
      \includegraphics[scale=0.13]{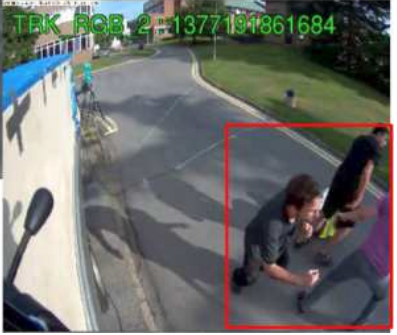}
    \end{minipage}\\ \hline

    \begin{tabular}[c]{@{}c@{}}PWPD~\cite{yi2015understanding} \\(2015)\end{tabular} &
    \begin{tabular}[c]{@{}c@{}}anomalous trajectory, \\abnormal crowd: at\\ terminal station\end{tabular}&
    \begin{tabular}[c]{@{}c@{}}trajectory prediction~\cite{fernando2018soft},\\
    anomaly detection~\cite{fernando2018soft},\\ 
    pedestrian speed detection~\cite{yi2015pedestrian}, \\
    crowd behaviour analysis~\cite{li2018deep}\end{tabular}& 
    \begin{minipage}{.2\textwidth}
      \includegraphics[scale=0.15]{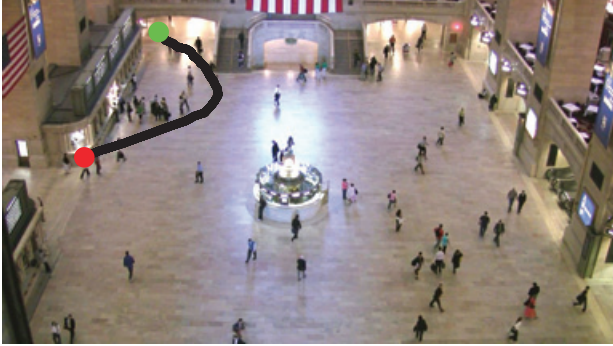}
    \end{minipage}&
    \begin{minipage}{.2\textwidth}
      \includegraphics[scale=0.15]{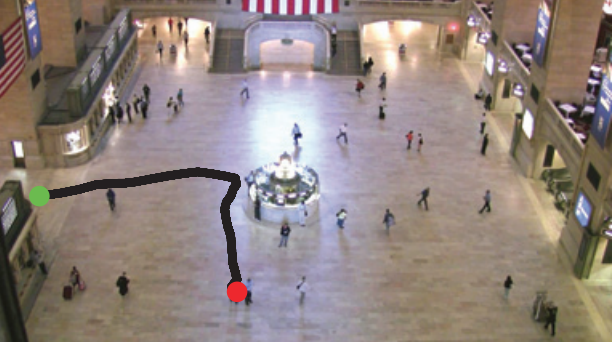}
    \end{minipage}\\ \hline
    
    \begin{tabular}[c]{@{}c@{}}RE-DID~\cite{rota2015real}\\(2015)\end{tabular} &
     \begin{tabular}[c]{@{}c@{}}fight: from \\multiple vehicle dash-cams\end{tabular}&
    fight detection~\cite{rota2015real}  &
    \begin{minipage}{.2\textwidth}
      \includegraphics[scale=0.045]{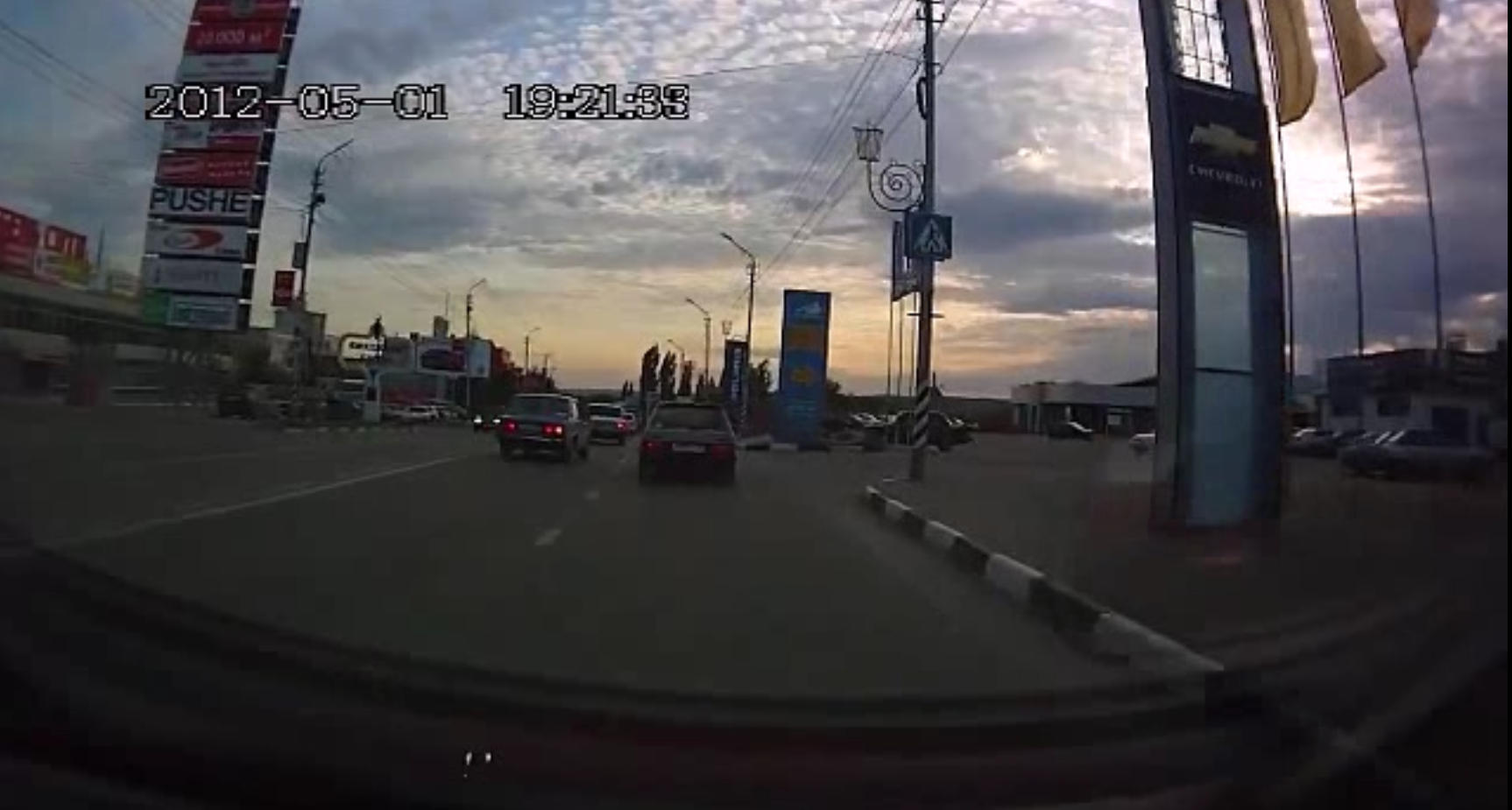}
    \end{minipage}&
    \begin{minipage}{.2\textwidth}
      \includegraphics[scale=0.2]{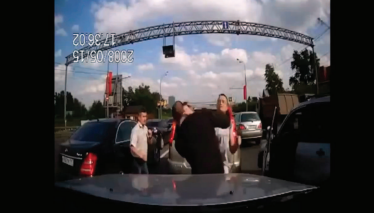}
    \end{minipage}\\ \hline
    
   \begin{tabular}[c]{@{}c@{}}MED~\cite{rabiee2016novel} \\(2016)\end{tabular}  &
   \begin{tabular}[c]{@{}c@{}}Panic, fight, congestion,\\ obstacle, neutral: \\at campus (walkway)\end{tabular}&
   \begin{tabular}[c]{@{}c@{}}anomaly detection~\cite{lazaridis2018abnormal},\\ panic detection~\cite{ammar2021deeprod}\end{tabular}&  
   \begin{minipage}{.2\textwidth}
      \includegraphics[scale=0.1]{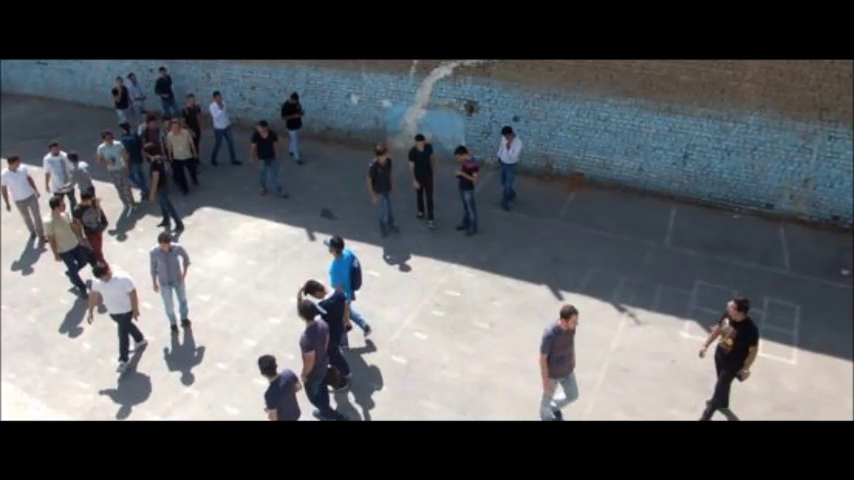}
    \end{minipage}&
    \begin{minipage}{.2\textwidth}
      \includegraphics[scale=0.1]{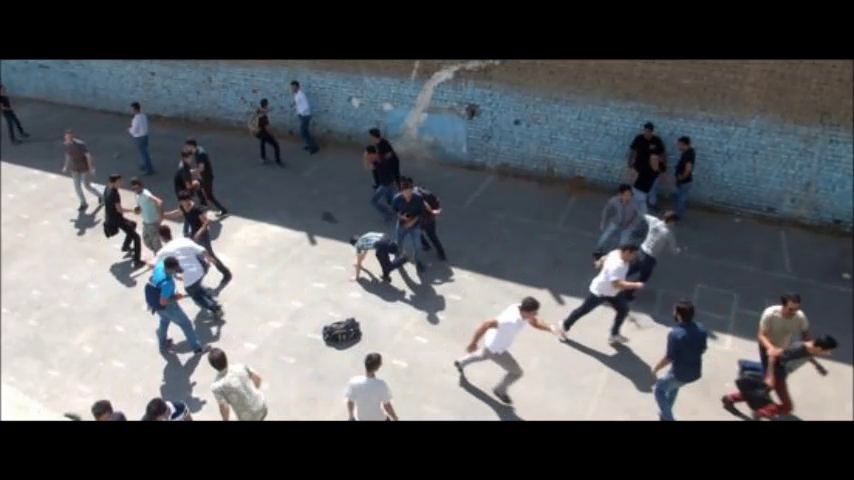}
    \end{minipage}\\ \hline

   \begin{tabular}[c]{@{}c@{}}ShanghaiTech~\cite{luo2017revisit}\\(2017)\end{tabular}  &
   \begin{tabular}[c]{@{}c@{}} bicycle, small carts,\\ fight, etc.: at\\ campus (ShanghaiTech) \end{tabular}&
    \begin{tabular}[c]{@{}c@{}}anomaly detection~\cite{gong2019memorizing,liu2018future}, \\
    action classification~\cite{zhong2019graph},\\
    crowd counting~\cite{sindagi2017cnn}\end{tabular}&  
     \begin{minipage}{.2\textwidth}
      \includegraphics[scale=0.1]{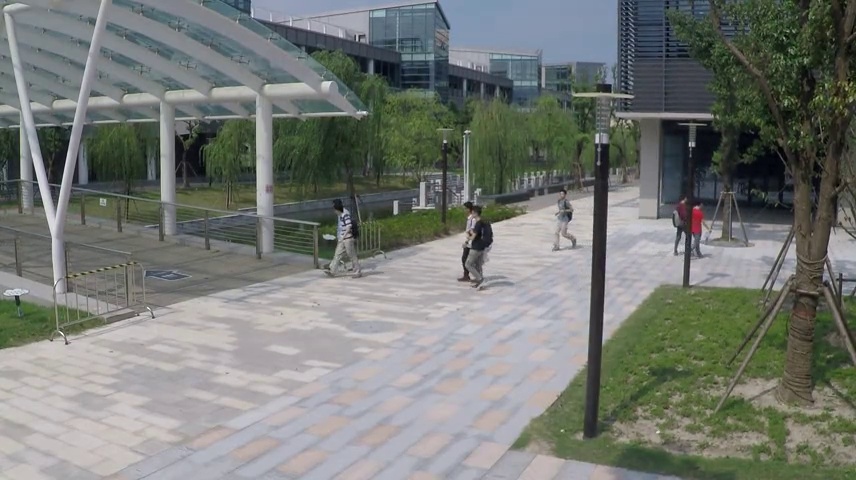}
    \end{minipage}&
    \begin{minipage}{.2\textwidth}
      \includegraphics[scale=0.1]{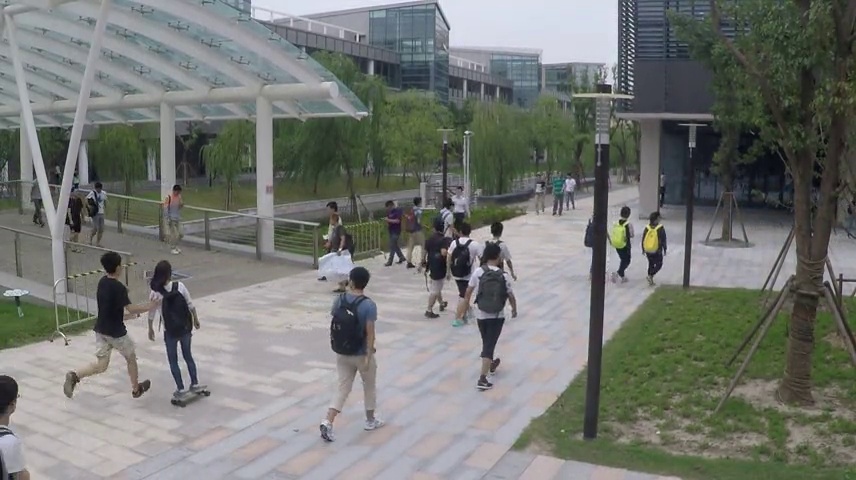}
    \end{minipage}\\ \hline
    
    \begin{tabular}[c]{@{}c@{}}LV~\cite{leyva2017lv}\\(2017)\end{tabular} &
      \begin{tabular}[c]{@{}c@{}}realistic security \\threats: at \\multiple location\end{tabular}& 
      \begin{tabular}[c]{@{}c@{}}
      anomaly detection~\cite{khan2018rejecting,leyva2017abnormal,deepak2021residual},\\
      panic detection~\cite{george2018crowd}\end{tabular}& 
    \begin{minipage}{.2\textwidth}
      \includegraphics[scale=0.18]{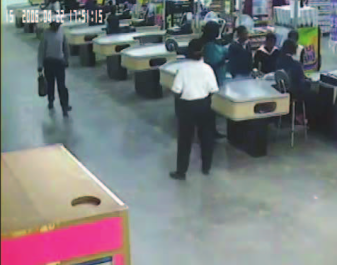}
    \end{minipage}&
    \begin{minipage}{.2\textwidth}
      \includegraphics[scale=0.18]{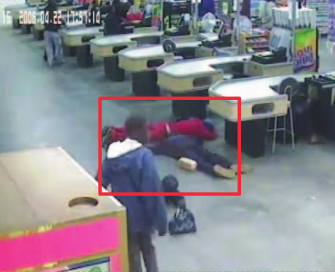}
    \end{minipage}\\ \hline
    
    \begin{tabular}[c]{@{}c@{}}UCF-Crime~\cite{sultani2018real}\\(2018)\end{tabular} &
    \begin{tabular}[c]{@{}c@{}} abuse, arrest,\\ assault, accident,\\ burglary, etc.: at \\multiple location\end{tabular} &
    \begin{tabular}[c]{@{}c@{}} anomaly detection~\cite{majhi2020temporal,ullah2021cnn},\\
    action classification~\cite{zhong2019graph}\end{tabular} & 
     \begin{minipage}{.2\textwidth}
      \includegraphics[scale=0.25]{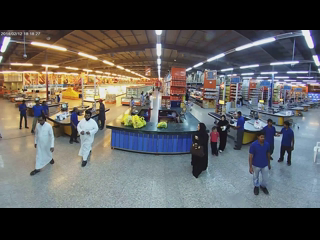}
    \end{minipage}&
    \begin{minipage}{.2\textwidth}
      \includegraphics[scale=0.25]{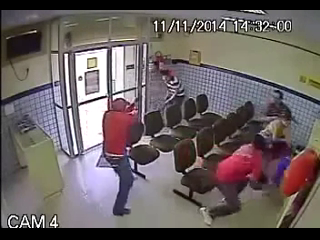}
    \end{minipage}\\ \hline

     \begin{tabular}[c]{@{}c@{}}IITH Accidents\\~\cite{singh2018deep}(2018)\end{tabular}  &
      \begin{tabular}[c]{@{}c@{}}accidents: at road\\ (intersection, junction)\end{tabular}&
    road accident detection~\cite{singh2018deep}&
    \begin{minipage}{.2\textwidth}
      \includegraphics[scale=0.4]{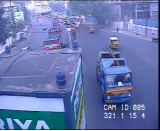}
    \end{minipage}&
    \begin{minipage}{.2\textwidth}
      \includegraphics[scale=0.4]{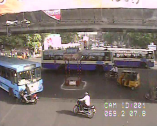}
    \end{minipage}\\ \hline
  
   \begin{tabular}[c]{@{}c@{}}CCTV-Fights\\~\cite{perez2019detection}(2019)\end{tabular} &
    \begin{tabular}[c]{@{}c@{}}fight: at \\multiple location\end{tabular} &
    fight detection~\cite{akti2019vision}&
  \begin{minipage}{.2\textwidth}
      \includegraphics[scale=0.06]{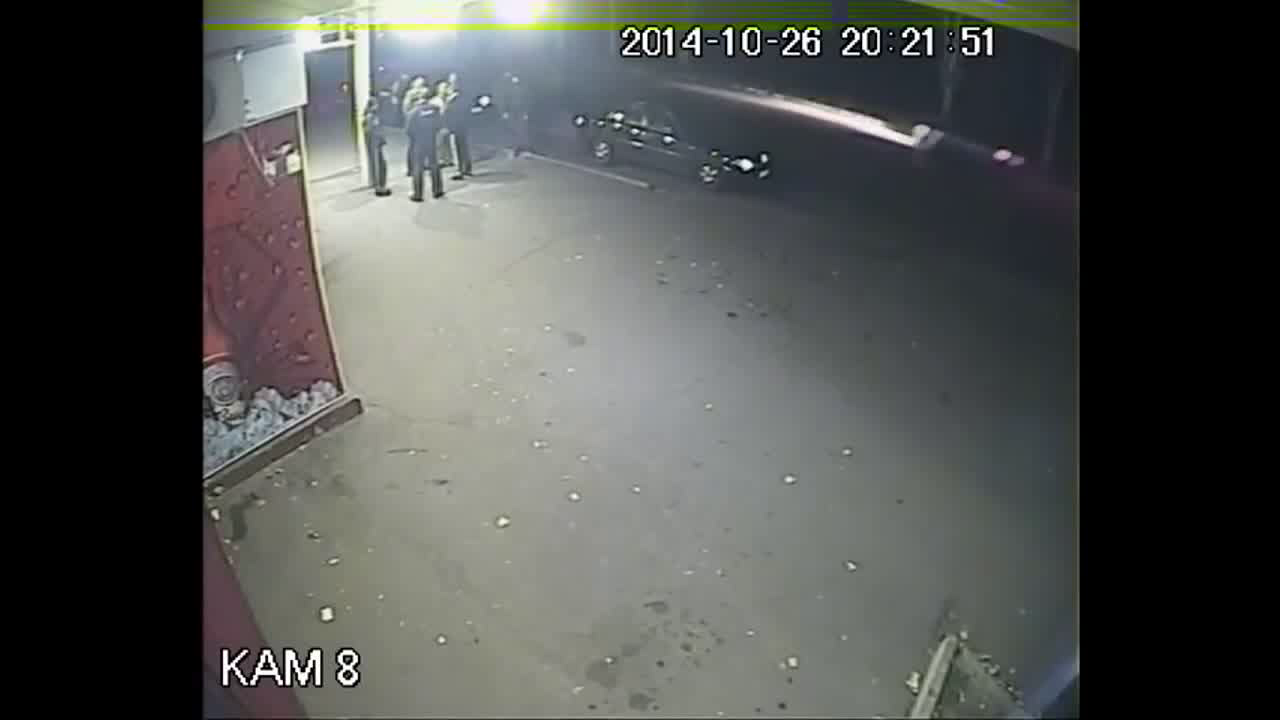}
    \end{minipage}&
    \begin{minipage}{.2\textwidth}
      \includegraphics[scale=0.06]{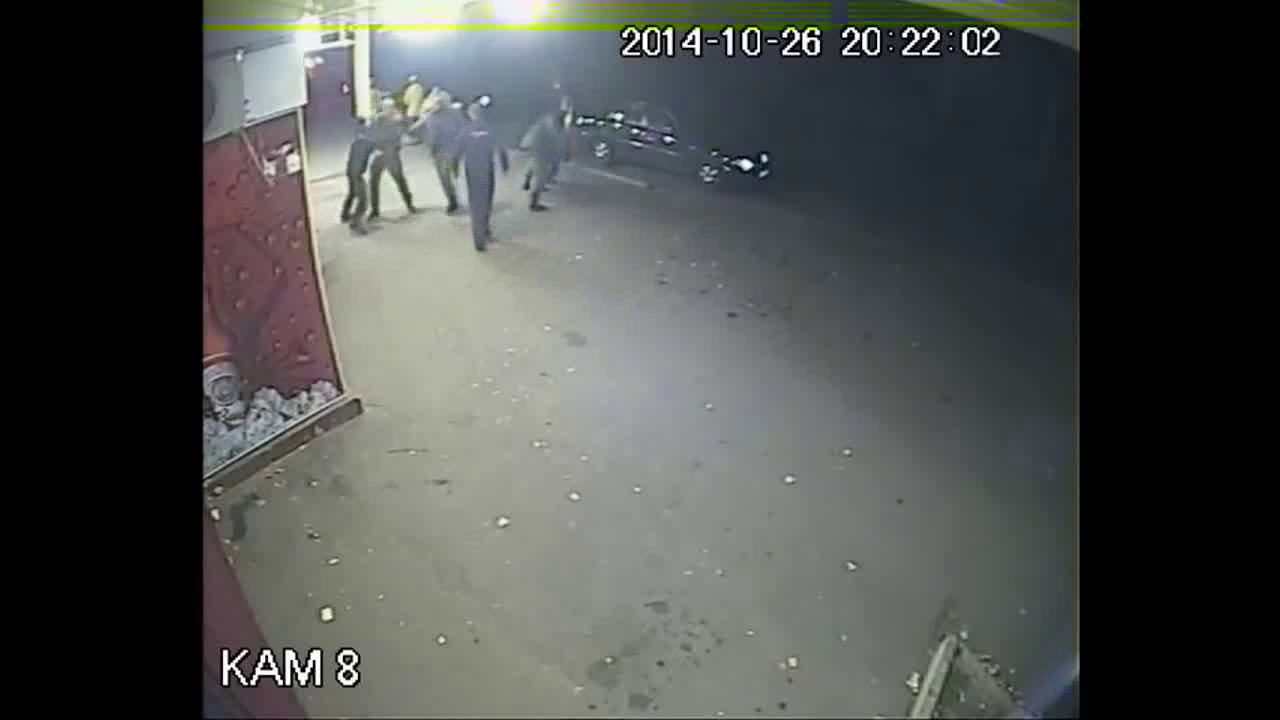}
    \end{minipage}\\ \hline

    \begin{tabular}[c]{@{}c@{}}Street Scene \\~\cite{ramachandra2020street}(2020)\end{tabular}  & 
    \begin{tabular}[c]{@{}c@{}} jaywalking across road, \\pedestrians loitering,\\ u-turns: on road (two-\\lane street)\end{tabular}&
    \begin{tabular}[c]{@{}c@{}}  anomaly detection~\cite{pourreza2021ano}, \\road surveillance~\cite{kapoor2020intelligent}\end{tabular}&  
     \begin{minipage}{.2\textwidth}
      \includegraphics[scale=0.06]{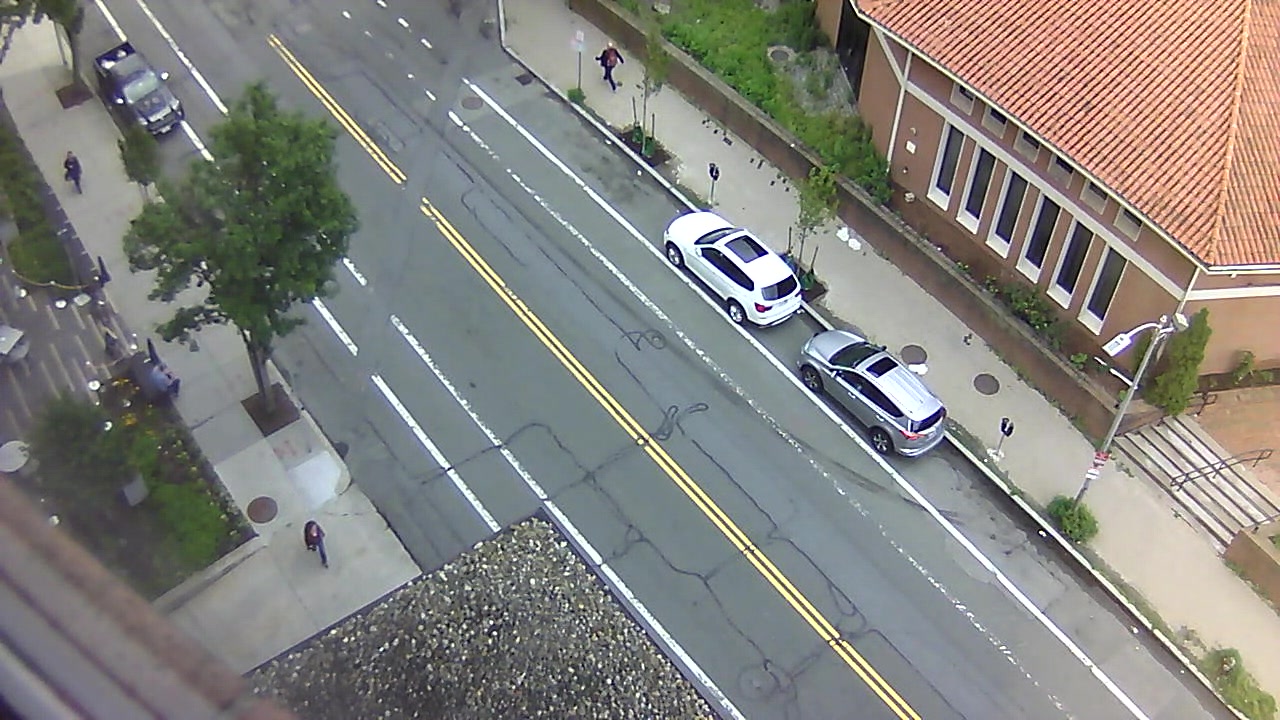}
    \end{minipage}&
    \begin{minipage}{.2\textwidth}
      \includegraphics[scale=0.06]{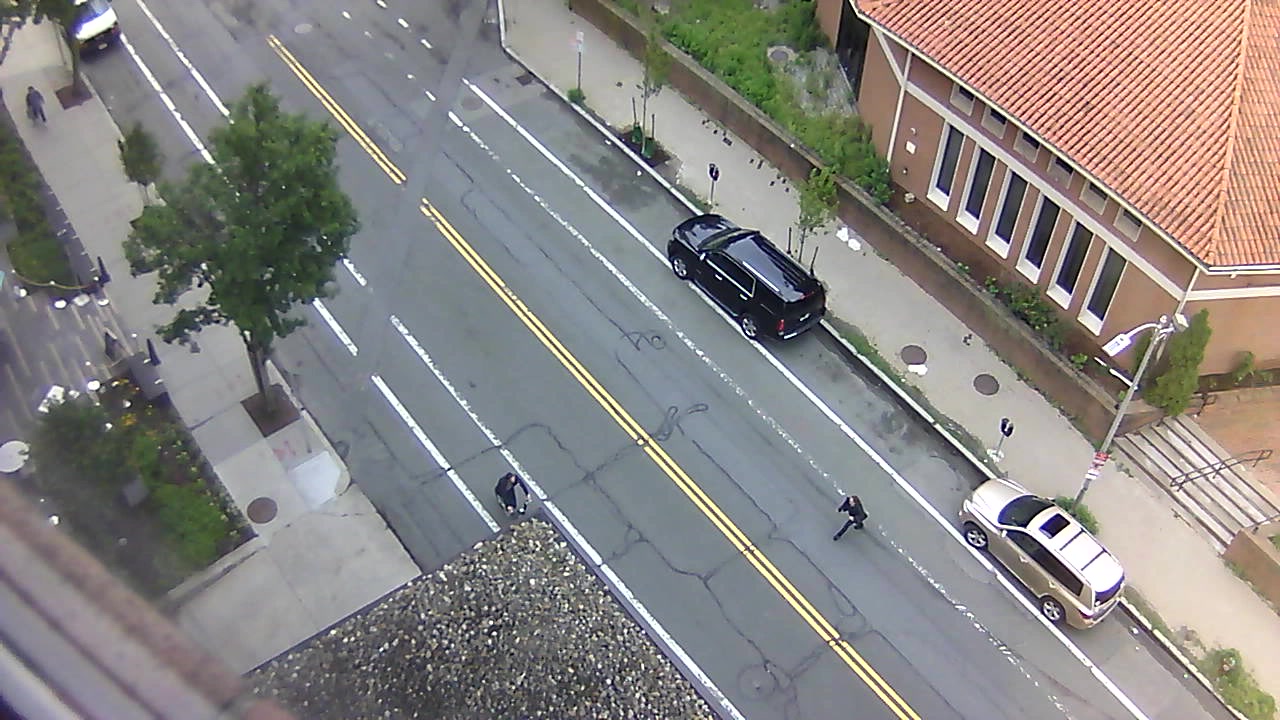}
    \end{minipage}\\ \hline

    \begin{tabular}[c]{@{}c@{}}ADOC~\cite{pranav2020day} \\(2020)\end{tabular}  &
    \begin{tabular}[c]{@{}c@{}}walking with balloons/dog,\\person on vehicle, \\crowd gathering, etc.:\\ at campus (walkway)\end{tabular}&
    anomaly detection~\cite{pourreza2021ano}& 
     \begin{minipage}{.2\textwidth}
      \includegraphics[scale=0.18]{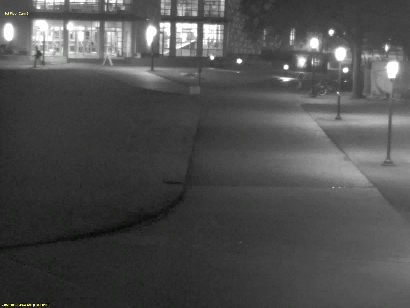}
    \end{minipage}&
    \begin{minipage}{.2\textwidth}
      \includegraphics[scale=0.18]{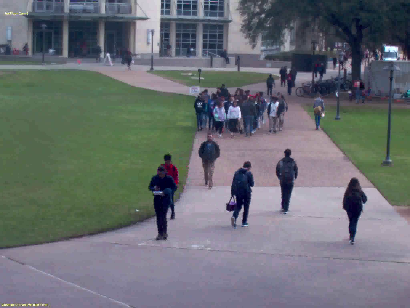}
    \end{minipage}\\ \hline
     
     \begin{tabular}[c]{@{}c@{}}HTA~\cite{singh2020anomalous} \\(2020)\end{tabular}  &
     \begin{tabular}[c]{@{}c@{}}accident, speeding \\vehicle, close merge:\\ at multiple location\end{tabular}&
     anomalous motion detection~\cite{singh2020anomalous} &  
      \begin{minipage}{.2\textwidth}
      \includegraphics[scale=0.06]{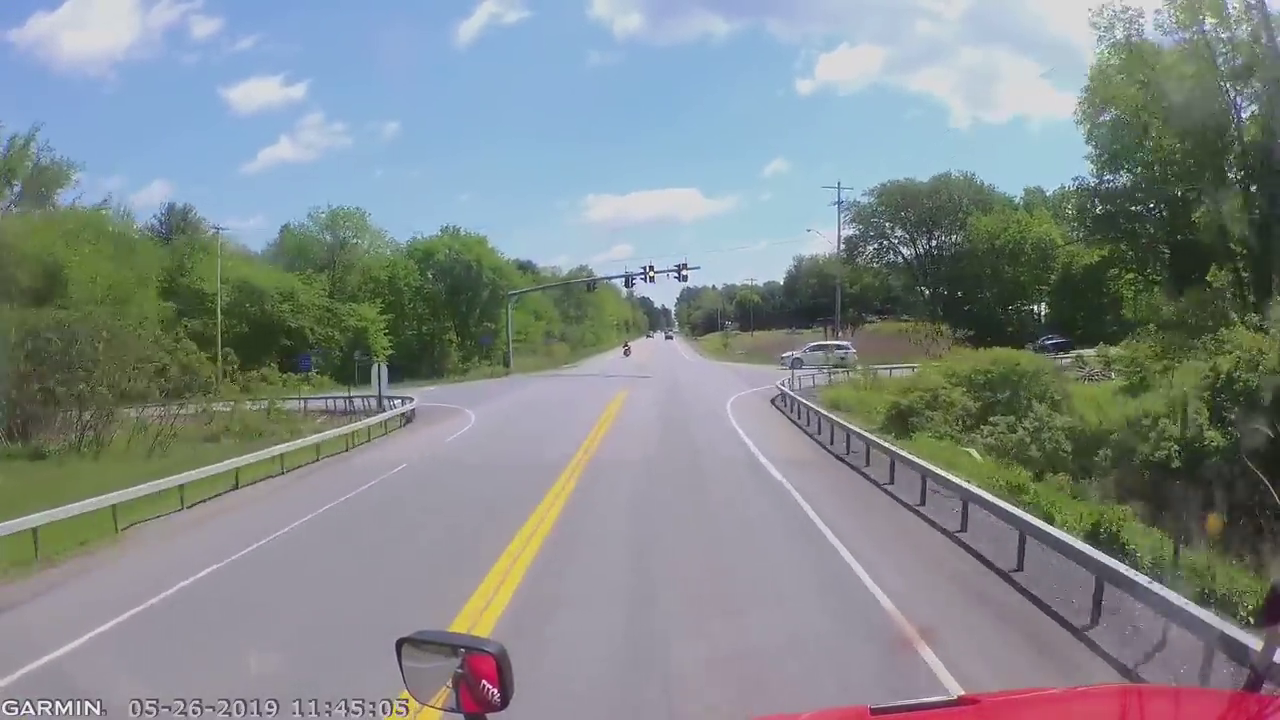}
    \end{minipage}&
    \begin{minipage}{.2\textwidth}
      \includegraphics[scale=0.06]{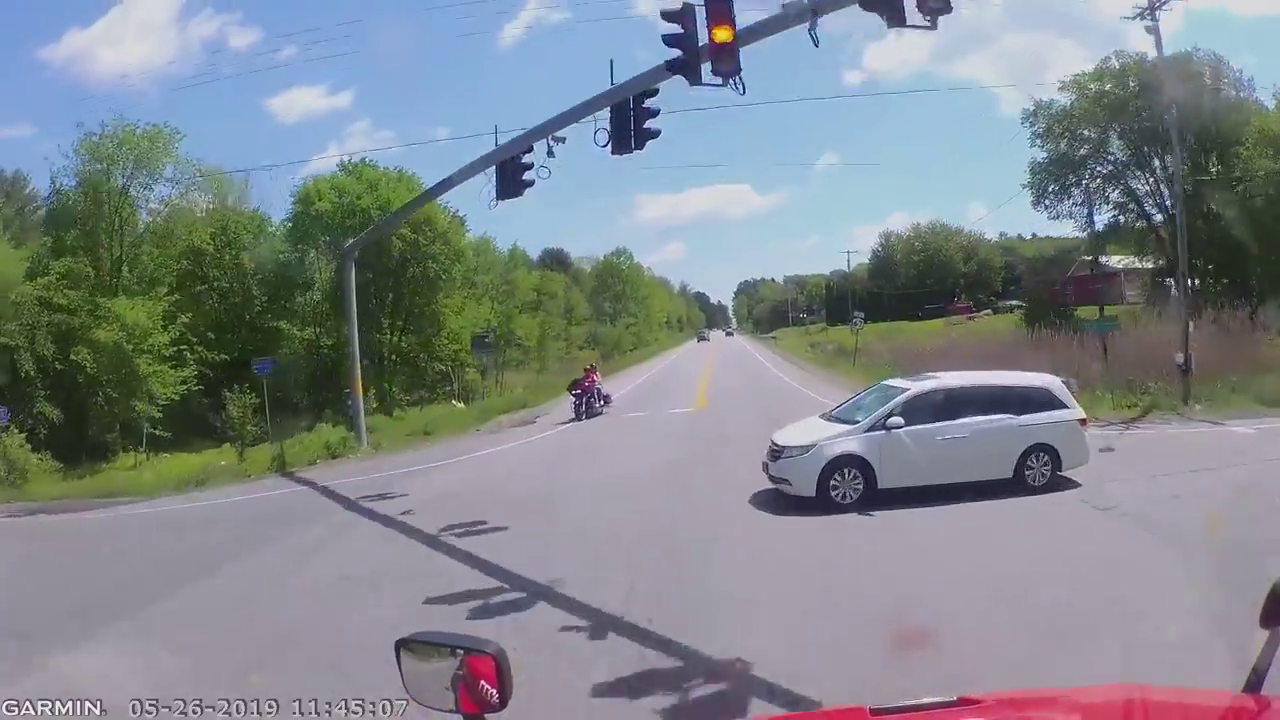}
    \end{minipage}\\ \hline

    %NVIDIA AI CITY&& accidents&& \\\hline
\end{tabular}}
\end{table*}
%===============================================================
%%%%%%%%%%%%%%%%%%%%%%%%%%%%%%%%%%%%%%%%%%%%%%%%%%%%%%

There are some other datasets that are not primarily developed for anomaly detection but other sub-tasks of scene monitoring such as detecting group formation, tracking people, counting people, human interaction/action classification, etc. We discuss them under the additional video dataset category in supplementary material attached with this survey, as they can also be utilized for generic scene surveillance or can be useful in creating some giant anomaly datasets.
%%%%%%%%%%%%%%%%%%%%%%%%%%%%%%%%%%%%%%%%%%%%%%%%%%%%%%%%%%%%%%%
%------------------------------------------------------------
\begin{table*}[!htbp]
\caption{Specifications of audio anomaly datasets}
\label{Tab:Specs}
\centering
 \resizebox{.99\textwidth}{!}{
 \begin{tabular}{|c|c|c|c|c|} \hline
Dataset &Continuity& Total duration  &No. of samples & Sampling rate \\ \hline

MIVIA  (2014)& no&3580 seconds & -&32 kHz \\\hline

AudioSet Soundscape: Ithaca, New York  & no&\begin{tabular}[c]{@{}c@{}} \#1: 797 hrs\\\#2: 638 hrs\end{tabular}  & - & \begin{tabular}[c]{@{}c@{}} 48 kHz\\ 48 kHz\end{tabular} \\

AudioSet Soundscape: Sabah, Malaysia&no&
\begin{tabular}[c]{@{}c@{}} Tuscam: 27 hrs 40 min\\Audiomoth: 784 hrs \end{tabular} & - &\begin{tabular}[c]{@{}c@{}} 44.1 kHz\\ 16 kHz\end{tabular} \\

AudioSet Soundscape: New Zealand (2015) &no& 240 hrs  &  -  & 32 kHz   \\
AudioSet Soundscape: Sulawesi  &no& 64 hrs    &  - & 48 kHz\\
AudioSet Soundscape: Republic of Congo & no&238 hrs 20 min  &  - & 8 kHz  \\\hline

DCASE 2017 (2017)& no&39 min  & 1170 & 44.1 kHz  \\\hline

ICBHI 2017 (2017)   &no& $>$ 5.5 hrs     & -& 4-44.1 kHz   \\\hline

SMD (2018)& no&$\approx$ 1 hour & 2048& 16.3 kHz \\\hline

MIMII DUE (2019)  &no& $>$420000 sec & 32157   &   16 kHz \\\hline

ToyADMOS (2019)&no& $\approx$ 540 hrs  & $>$ 12000&   48 kHz  \\\hline

LIFE DYNAMAP (2020)  &no& 550 sec  & - &  48 kHz \\\hline

ToyADMOS2 (2021)&no& 604 hrs & $\approx$264k &  48 kHz\\ \hline

DCASE 2021 (2021)  &no& $\approx$82 hours & 29463 & 16 kHz\\\hline
\end{tabular}
}
\end{table*}
%--------------------------------
%%%%%%%%%%%%%%%%%%%%%%%%%%%%%%%%%%%%%%%%%%%%
\section{Audio anomaly datasets}\label{sec:audioDatasets}
Audio anomaly detection is gaining a lot of attention in many applications in various fields, such as traffic surveillance, industries, music, medicines, etc. Unlike for video, not many audio anomaly datasets are available for study. A few datasets have been created based on applications and further provided publicly. In this section, we describe and discuss the features of publicly available audio datasets.
%%%%%%%%%%%%%%%%%%%%%%%%%%%%%%%%%%%%%%%%%%%%%%%%%%%%%%%%%%
%%%%%%%%%%%%%%%%%%%%%%%%%%%%%%%%%%%%%%%%%%%%%%%%%%%
%\subsection{Dataset specifications}
Table~\ref{Tab:Specs} gives the specifications of the audio samples collected for different datasets. The specifications include time duration of each dataset, number of samples collected, and sampling rate for audio-digitization. It can be observed from the table that the duration of audio for different datasets varies from a few seconds to a number of hours. AudioSet soundscape dataset has the longest duration of audios, followed by ToyADMOS2~\cite{harada2021toyadmos2} and ToyADMOS~\cite{koizumi2019toyadmos} datasets. A very small recording of 550 seconds has been collected for the LIFE DYNAMAP project~\cite{socoro2015development}.

Further, from the table, it can be seen that the sampling rate for different audio datasets varies between the range of 4kHz to 48 kHz. The sampling rate of 48 kHz has been mostly considered for audio digitization in various datasets. Also, for ICBHI 2017~\cite{pham2020robust}, there is no fixed value of sampling rate for the data; rather, it varies between the range of 4 to 44.1 kHz. Thus, different sampling rates have been used as per the requirement of the tasks.

%\subsection{Dataset classification}
Dataset classification based on various attribute as mentioned in Fig.~\ref{fig:taxonomy} is discussed in this section. Unlike in video datasets, since only segment level labelling is followed for the audio datasets, hence, a separate classification of audio datasets based on labelling has not been provided.
%-------------------------------
\begin{figure*}[!htbp]
  \centering
    \includegraphics[scale=0.4]{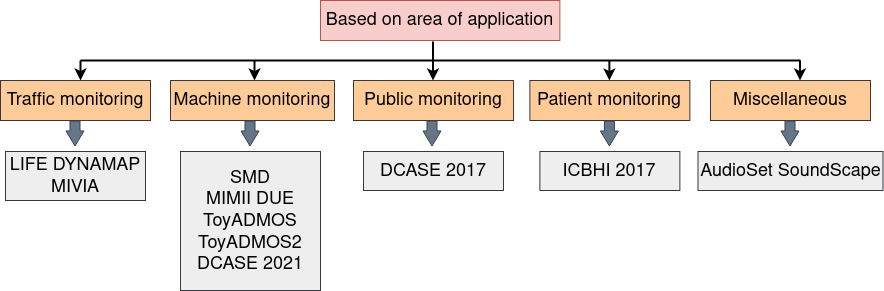}
  \caption{Categorization of audio anomaly datasets based on area of application}
  \label{fig:audio_taxonomy_application_based}
\end{figure*}
%-------------------------------
%-------------------------------
\begin{figure}[!htbp]
  \centering
    \includegraphics[scale=0.4]{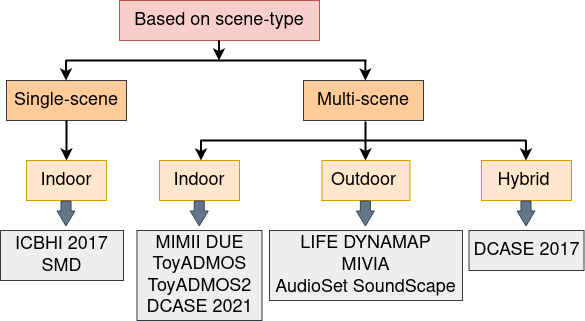}
   \caption{Categorization of audio anomaly datasets based on scene-type}
  \label{fig:audio_taxonomy_sceneType_based}
\end{figure}
%-------------------------------
%-------------------------------
\begin{figure}[!htbp]
  \centering
    \includegraphics[scale=0.4]{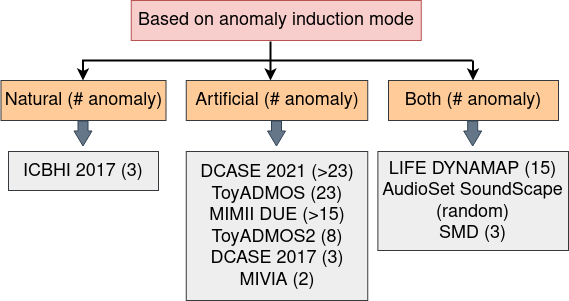}
  \caption{Categorization of audio anomaly datasets based on anomaly induction mode}
  \label{fig:audio_taxonomy_induction_based}
\end{figure}
%-------------------------------
% %-------------------------------
% \begin{figure}[!htbp]
%   \centering
%     \includegraphics[scale=0.6]{DRAWN/audio_taxonomy_labeling_based.png}
%   \caption{Categorization of audio anomaly datasets based on labeling procedure}
%   \label{fig:audio_taxonomy_labeling_based}
% \end{figure}
% %-------------------------------
%%%%%%%%%%%%%%%%%%%%%%%%%%%%%%%%%%%%
\subsection{Area of application}
Audio anomaly datasets cover a variety of applications such as monitoring of traffic, machine, public, patient, etc. The area of application based classification is shown in Fig.~\ref{fig:audio_taxonomy_application_based}. Majority of the audio anomaly datasets have been developed for machine monitoring, whereas a few datasets are available for monitoring traffic and public. Heath supervision is also made possible using ICBHI~\cite{pham2020robust} dataset by monitoring respiratory patterns in the patients.   
%%%%%%%%%%%%%%%%%%%%%%%%%%%%%%%%%%%%
\subsection{Scene-type}
Scene based categorization for audio datasets is shown via Fig.~\ref{fig:audio_taxonomy_sceneType_based}. Owing to different recording locations for each dataset, the number of multi-scene datasets is much higher and more easily available as compared to those collected from one location, such as ICBHI~\cite{pham2020robust} and SMD~\cite{oh2018residual}. The single-scene datasets have been collected in indoor scenarios, whereas multi-scene datasets have been created at various indoor and outdoor locations such as industry, road, forests, etc.
%%%The datasets have been created at various indoor and outdoor locations, such as industry, roads, forests, etc.
%%%%%%%%%%%%%%%%%%%%%%%%%%%%%%%%%%%%%%
%%%%%%%%%%%%%%%%%%%%%%%%%%%%%%%%%%%%%%%%%%%%%%%%%%%%%%%%%%
%%%%%%%%%%%%%%%%%%%%%%%%%%%%%%%%%%%%
\subsection{Anomaly induction mode}
Fig.~\ref{fig:audio_taxonomy_induction_based} shows anomaly induction mode based categorization of datasets. The count of distinct anomalies present in the dataset is also listed along with. As it can be seen from the figure, most of the audio datasets have been created artificially, except for ICBHI 2017~\cite{pham2020robust} dataset, which records the natural respiratory sounds from a human body. 
%%%The datasets have been created at various indoor and outdoor locations, such as industry, roads, forests, etc. 
Since it is difficult to generate natural audio datasets consisting of anomalies, hence anomalies have been generated deliberately to simulate the system for its detection and for future anomaly detection applications. 
%%%%%%%%%%%%%%%%%%%%%%%%%%%%%%%%%%%%%%%%%%%%%%%%%%%%%%%%%%%%%%%
% \subsection{Labelling procedure}
% ???Fig.~\ref{fig:audio_taxonomy_labeling_based} shows labelling procedure based categorization of datasets.
%%%%%%%%%%%%%%%%%%%%%%%%%%%%%%%%%%%%%%%%%%%%%%%%
%%%%%%%%%%%%%%%%%%%%%%%%%%%%%%%%%%%%%%%%%%%%%%%%%%%%%%%%%%%%555
\subsection{Dataset Overview}
Table~\ref{tab:audioTableMain} list all the publicly available audio anomaly datasets in chronological order of their release year. Some of the anomalies that have been generated are mentioned for every dataset in the table. The datasets have been used for applications in different fields, such as in industries for machine monitoring and inspection, road surveillance, etc. A dataset for anomaly detection in respiratory patterns in human bodies has also been created. Most of the datasets have been artificially generated by deliberately adding anomalies and made available to the public recently.
%=====================================================
\begin{table*}[!htbp]
  \caption{Audio anomaly datasets: primary information}
  \label{tab:audioTableMain}
  \centering
  \resizebox{.99\textwidth}{!}{
  \begin{tabular}{|c|c|c|}
    \hline
     Dataset &  Anomalies: collection scenario & Application\\\hline
    
      \begin{tabular}[c]{@{}c@{}}MIVIA~\cite{foggia2014cascade} \\(2014)\end{tabular}    & \begin{tabular}[c]{@{}c@{}}  \\   \end{tabular} crashes and tire skidding: on road &road surveillance~\cite{strisciuglio2019learning,almaadeed2018automatic,greco2020aren}  \\ \hline

\begin{tabular}[c]{@{}c@{}}AudioSet Soundscape~\cite{sethi2020characterizing}\\(2015)\end{tabular} & \begin{tabular}[c]{@{}c@{}}music, human speech, machine\\ noise, etc.: from nature   \end{tabular}&   
\begin{tabular}[c]{@{}c@{}}anomaly detection from \\eco-acoustic data \cite{sethi2020characterizing} \end{tabular}\\\hline

\begin{tabular}[c]{@{}c@{}}DCASE 2017~\cite{DCASE2017}\\(2017)\end{tabular}  &
\begin{tabular}[c]{@{}c@{}} baby cry, glass break, \\gun shot: city   \end{tabular}  & detection of rare sounds \cite{provotar2019unsupervised,rushe2019anomaly}  \\\hline

\begin{tabular}[c]{@{}c@{}}ICBHI 2017~\cite{pham2020robust}\\(2017)\end{tabular}& 
 \begin{tabular}[c]{@{}c@{}}crackle, wheeze, crackle and\\ wheeze: in human body  \end{tabular}&  \begin{tabular}[c]{@{}c@{}}anomalous respiratory pattern\\ detection in humans \cite{chen2019triple,acharya2020deep,demir2020convolutional}\end{tabular}\\\hline

\begin{tabular}[c]{@{}c@{}}SMD~\cite{oh2018residual}\\(2018)\end{tabular} & 
 \begin{tabular}[c]{@{}c@{}} non-greased line, B-line, C-line\\ to B-line: from industry \end{tabular} & machine monitoring in industries \cite{oh2018residual}  \\\hline

\begin{tabular}[c]{@{}c@{}}MIMII DUE~\cite{tanabe2021mimii}\\(2019)\end{tabular}  
& \begin{tabular}[c]{@{}c@{}} wing damage, clogging, gear, \\contamination etc.: from industry \end{tabular}  &  \begin{tabular}[c]{@{}c@{}} investigation and inspection\\ of industrial machine\end{tabular}\\\hline

\begin{tabular}[c]{@{}c@{}}ToyADMOS~\cite{koizumi2019toyadmos}\\(2019)\end{tabular}  &  \begin{tabular}[c]{@{}c@{}}deformed gears, over/under voltage,\\ pulley,chipped wheel, axle, excessive\\ tension, etc.: from industry   \end{tabular}& \begin{tabular}[c]{@{}c@{}} anomalous sound detection in \\miniature machines \cite{primus2020anomalous,koizumi2020spidernet} \end{tabular}  \\\hline

\begin{tabular}[c]{@{}c@{}}LIFE DYNAMAP~\cite{socoro2015development}\\(2020)\end{tabular}  &
 \begin{tabular}[c]{@{}c@{}}horns, church bells, birds,\\ thunder etc.: on  road \end{tabular}& 
 \begin{tabular}[c]{@{}c@{}} detection and removal of anomalous\\ events for road traffic mapping \cite{socoro2017anomalous,alias2017description}\end{tabular} \\\hline

\begin{tabular}[c]{@{}c@{}}ToyADMOS2~\cite{harada2021toyadmos2}\\(2021)\end{tabular}   &
 \begin{tabular}[c]{@{}c@{}}bent shaft, melted gears,\\ flat tire etc.: from industry\end{tabular}& \begin{tabular}[c]{@{}c@{}}anomalous sound detection \\in miniature machines \end{tabular}\\\hline

\begin{tabular}[c]{@{}c@{}}DCASE 2021~\cite{kawaguchi2021description} \\(2021)\end{tabular}   &
 \begin{tabular}[c]{@{}c@{}}wing damage, clogging, chipped\\ wheel, axle etc.: from industry \end{tabular}  & 
 \begin{tabular}[c]{@{}c@{}}detection of anomalous sounds during\\ machine monitoring \cite{morita2021anomalous,wilkinghoff2021utilizing,kuroyanagi2021anomalous} \end{tabular} \\
\hline
\end{tabular}
}
\end{table*}
%===============================================================
%=====================================================
\begin{table*}[!htbp]
  \caption{Specifications of audio-visual anomaly datasets}
  \label{tab:specification_av}
  \resizebox{.99\textwidth}{!}{
  \begin{tabular}{|c|c|c|c|c|c|c|c|}
    \hline
Dataset&Continuity& \begin{tabular}[c]{@{}c@{}} Total\\Duration\end{tabular}& \begin{tabular}[c]{@{}c@{}} Total No. of\\Frames\end{tabular}   &  \begin{tabular}[c]{@{}c@{}} Total No. of\\Videos\end{tabular}  &
\begin{tabular}[c]{@{}c@{}} Video\\resolution\end{tabular}  &FPS (video) & \begin{tabular}[c]{@{}c@{}} Camera\\motion\end{tabular}  \\\hline
\begin{tabular}[c]{@{}c@{}} Human-human \\interaction (2014)\end{tabular} &no& 32.24 min&- &8 &-& -&none\\
VSD (2015)&no&35 hrs 18 min& -& 25&multiple&multiple&in some \\
EMOLY (2018)&no& -    &-&123&-&-&none\\
XD-Violence (2020) &no& 217 hrs&-&4754&multiple&multiple&in some \\
BAREM (2021)&no& $\approx$ 6 hrs&$\approx$5,40,000&72&-&25&none \\
\hline
\end{tabular}
}
\end{table*}
%===============================================================
%%%%%%%%%%%%%%%%%%%%%%%%%%%%%%%%%%%%%%%%%%%%%%%%%%%%%%%%%%%
\section{Audio-visual datasets}\label{sec:audioVisual}
So far, we have discussed a number of publicly available datasets for video and audio anomaly detection applications individually. Since a course of actions consists of both audio and video components dependent on each other, hence, audio-visual analysis can result in more accurate results compared to audio and video recordings being used individually~\cite{kumari2021situational,atrey2010multimodal}. 
There are very few audio-visual datasets for anomaly detection. They are mainly developed for applications such as detection of anomalous expression, violence, stress, etc. %Now, we will discuss the specification and categorization of audio-visual datasets.

%%Some of the application include anomalous expression detection, violence detection, detection of stress in a particular situation, etc.
%%%%%%%%%%%%%%%%%%%%%%%%%%%%%%%%%%%%%%%%%%%%%%%%%%%%%%%%%%
%%%%%%%%%%%%%%%%%%%%%%%%%%%%%%%%%%%%
%\subsection{Dataset specifications}
%%%\hl{Datasets specification of all the five datasets are given in Table{~\ref{tab:audioVisualTableBig4}}. The table enables a quick comparison of datasets based on their total frame count, resolution, duration, etc.}
Table~\ref{tab:specification_av} provides details about the features and specifications of the audio-visual clips. The table enables a quick comparison of datasets based on their total frame count, resolution, duration, etc. Mainly, information about the total number of videos and total duration is made available by the authors. The datasets for violence detection records the largest length of 217 hours compared to others. This is because violence event is comparatively more often to occur and hence easily available in CCTV footage. The total number of videos available for XD-Violence~\cite{wu2020not} is maximum, i.e., 4754, followed by the EMOLY~\cite{fayet2018emo} dataset consisting of 123 videos. Least number of videos have been provided in the Human-human interaction dataset, with a total duration of 32.24 minutes. Clips have been recorded at multiple resolutions and at multiple frame rates. However, the sampling rate of audios recorded in the databases has not been made available by the authors. Few clips from VSD~\cite{demarty2015vsd} as well as XD-Violence~\cite{wu2020not} datasets consist of camera motion, thus making them challenging for further analysis.

\begin{figure}[!htbp]
  \centering
    \includegraphics[scale=0.45]{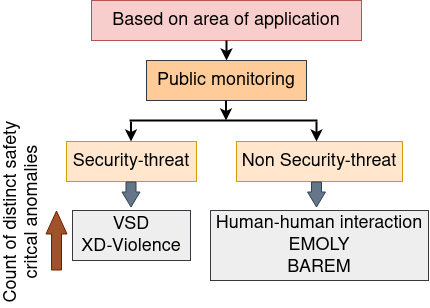}
   \caption{Categorization of audio-visual anomaly datasets based on area of application}
  \label{fig:audioVideo_taxonomy_application_based}
\end{figure}

\begin{figure}[!htbp]
  \centering
    \includegraphics[scale=0.45]{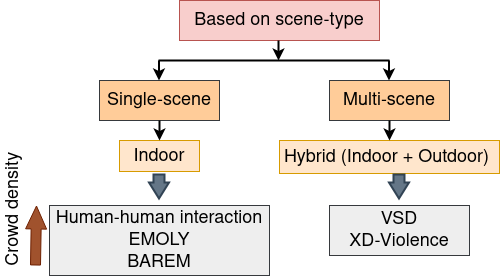}
   \caption{Categorization of audio-visual anomaly datasets based on scene-type}
  \label{fig:audioVideo_taxonomy_sceneType_based}
\end{figure}

Also, it can be observed from the table that none of the datasets is collected in continuation and hence are not suitable for evaluation of adaptive anomaly detection frameworks.
%%%%%%%%%%%%%%%%%%%%%%%%%%%%%%%%%%%%%%%%%%%%%%%%%%%%%%%%%%%%%%%%%%%%%%%%%%%%%%%%%%%%
%%%%%%%%%%%%%%%%%%%%%%%%%%%%%%%%%%%%%%%%%%%%%%%5
%\subsection{Dataset classification}
On the same categorization basis as for audio datasets, we present and discuss the categorization of audio-visual datasets through Fig.~\ref{fig:audioVideo_taxonomy_application_based} to Fig.~\ref{fig:audioVideoo_taxonomy_induction_based}.
Unlike in video datasets, since only frame level labelling is followed for the audio-visual datasets, hence, a separate class of audio-visual datasets based on labelling has not been provided.
Since all the datasets possess frame-level annotations, hence, anomaly localization is not yet explored in audio-visual datasets.
%%%%%%%%%%%%%%%%%%%%%%%%%%%%%%%%%%%%%%%%%%%%%%%%%%%%%%%%%%%%%%%%%%%%%%%%
%-------------------------------
%-------------------------------
\subsection{Area of application}
The audio-visual datasets have been designed and made available to the users recently. They all fall into public monitoring applications. The categorization is shown in Fig.~\ref{fig:audioVideo_taxonomy_application_based}. Different sub-applications of public monitoring, such as anomalous expression detection, violence detection, detection of stress in a particular situation, etc., have been covered for these datasets. 
Based on whether the anomalies posses security threats or not, they are further categorized into two groups viz., security-threat and non security-threat.
%. The categorization is shown in Fig.~\ref{fig:audioVideo_taxonomy_application_based}.
VSD~\cite{demarty2015vsd} and XD-Violence~\cite{wu2020not} datasets have security threat-based anomalies, and the rest datasets do not possess any security threat-based anomalies.
%%%We further categorize the datasets based on nature of anomalies, i.e., whether the present anomalies are security threat or non-security threat. 
%%%%%%%%%%%%%%%%%%%%%%%%%%%%%%%%%%%%
%-------------------------------
%-------------------------------
%%%%%%%%%%%%%%%%%%%%%%%%%%%%%%%%%%%%%%%%%%%%%%
\subsection{Scene-type}
Scene-type-based dataset categorization is shown via Fig.~\ref{fig:audioVideo_taxonomy_sceneType_based}. The audio-visual dataset collection scenarios are mostly limited to indoor. Human-human interaction~\cite{lefter2014audio}, EMOLY~\cite{fayet2018emo}, and BAREM~\cite{belmonte2021barem} are single-scene datasets, all collected in indoor environments.  %%The dataset collection scenarios are mostly limited to indoor, except for VSD and XD-Violence, which consist of both indoor and outdoor environments. 
Rest two datasets, viz., VSD~\cite{demarty2015vsd} and XD-Violence~\cite{wu2020not}, are multiscene datasets comprising of both indoor as well as the outdoor scenes. The datasets for violence detection, i.e., VSD~\cite{demarty2015vsd} and XD-Violence~\cite{wu2020not}, are mainly collected from real CCTV footage and hence have high crowd density; whereas the other datasets mainly contain actors and thus have low density. Among the three single-scene datasets, BAREM~\cite{belmonte2021barem} has the least crowd density (single person at a time), followed by EMOLY~\cite{fayet2018emo} and then Human-human interaction~\cite{lefter2014audio} dataset.
%%%%%%%%%%%%%%%%%%%%%%%%%%%%%%%%%%%%
%-------------------------------
\begin{figure}[!htbp]
  \centering
    \includegraphics[scale=0.4]{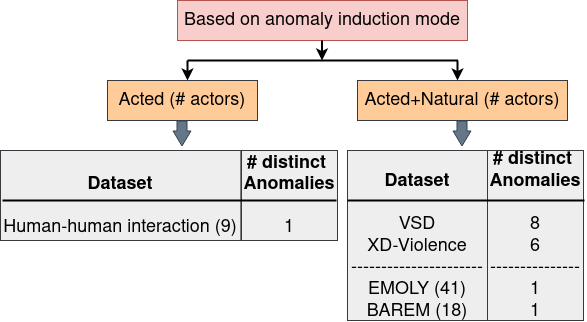}
     \caption{Categorization of audio-visual anomaly datasets based on anomaly induction mode}
  \label{fig:audioVideoo_taxonomy_induction_based}
\end{figure}
%-------------------------------
\subsection{Anomaly induction mode}
Categorization based on anomaly induction mode, i.e., natural or acted, is shown via Fig.~\ref{fig:audioVideoo_taxonomy_induction_based}. All the datasets contain some acted anomalies too. The number of actors in the datasets is also mentioned in parenthesis. Each dataset has only one type of distinct anomaly except for the violence detection datasets, i.e., VSD~\cite{demarty2015vsd} and XD-Violence~\cite{wu2020not}. VSD~\cite{demarty2015vsd} consists of 8 distinct anomalies, while XD-Violence~\cite{wu2020not} contains 6 distinct anomalies. Some of the example anomalies in audio-visual datasets include fights, stress, explosions, abuse, car accident, shooting, frustration, etc.
%%%%%%%%%%%%%%%%%%%%%%%%%%%%%%%%%%%%
%=====================================================
\begin{table*}[!htbp]
  \caption{Audio-visual anomaly datasets: primary information}
  \label{tab:audioVideoTableMain}
  \centering
  \resizebox{.99\textwidth}{!}{
  \begin{tabular}{|c|c|c|c|c|}
    \hline
     Dataset &  \begin{tabular}[c]{@{}c@{}}Anomalies: \\collection scenario \end{tabular}&Application &Anomaly image example&Normal image example\\\hline
    
     \begin{tabular}[c]{@{}c@{}} Human-human \\interaction~\cite{lefter2014audio} \\(2014)\end{tabular} &
      \begin{tabular}[c]{@{}c@{}}stress: at\\ help-desk \end{tabular} &  
       \begin{tabular}[c]{@{}c@{}}Detection of stressful \\situations at a\\ help-desk~\cite{lefter2015recognizing}\end{tabular}&
     \begin{minipage}{.2\textwidth}
      \includegraphics[scale=0.23]{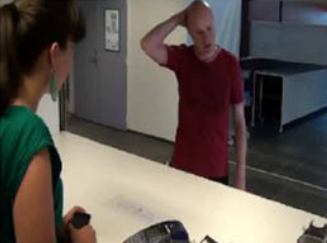}
    \end{minipage}&
    \begin{minipage}{.2\textwidth}
      \includegraphics[scale=0.23]{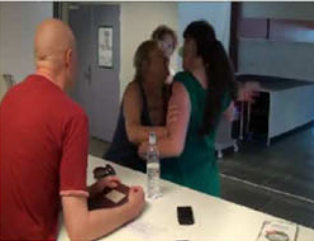}
    \end{minipage}\\ \hline

     \begin{tabular}[c]{@{}c@{}} VSD~\cite{demarty2015vsd}\\(2015)\end{tabular}&
     \begin{tabular}[c]{@{}c@{}}fights, fire, gunshot,\\ cold weapons, \\car chases, gory, \\exploision, screams: \\at multiple location\end{tabular}& 
      \begin{tabular}[c]{@{}c@{}}violence \\detection~\cite{peixoto2021harnessing,li2016detecting,khan2019cover}\end{tabular}&
     \begin{minipage}{.2\textwidth}
      \includegraphics[scale=0.15]{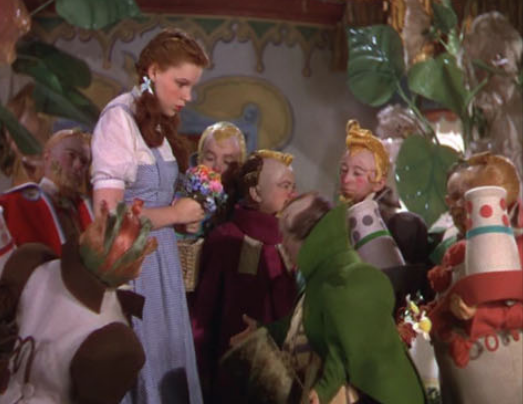}
    \end{minipage}&
    \begin{minipage}{.2\textwidth}
      \includegraphics[scale=0.15]{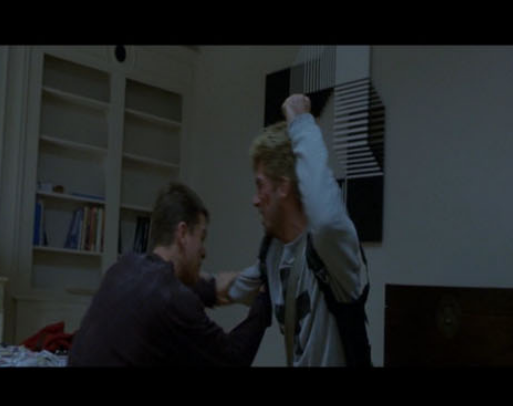}
    \end{minipage}\\ \hline
     
    \begin{tabular}[c]{@{}c@{}} EMOLY~\cite{fayet2018emo}\\(2018)\end{tabular}  &
    \begin{tabular}[c]{@{}c@{}} anomalous \\expression: in lab\end{tabular}& 
     \begin{tabular}[c]{@{}c@{}}abnormal expression \\detection~\cite{fayet2018emo}\end{tabular}&
     \begin{minipage}{.2\textwidth}
      \includegraphics[scale=0.15]{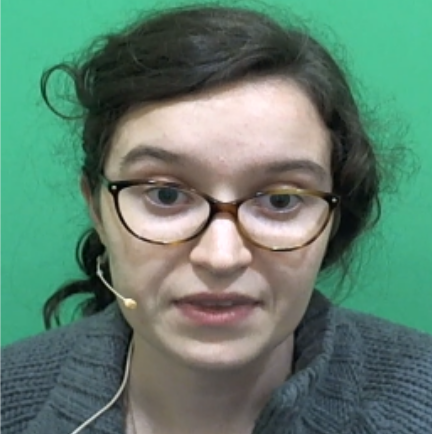}
    \end{minipage}&
    \begin{minipage}{.2\textwidth}
      \includegraphics[scale=0.15]{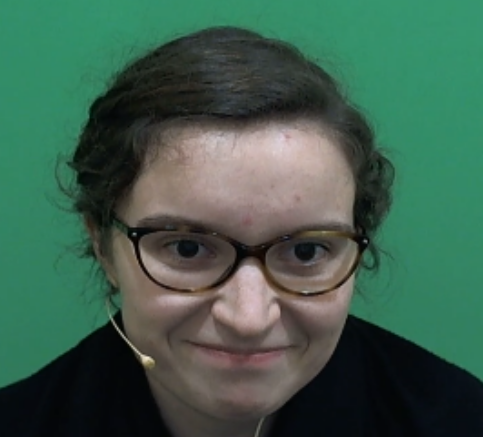}
    \end{minipage}\\ \hline
    
    \begin{tabular}[c]{@{}c@{}} XD-Violence~\cite{wu2020not}\\(2020)\end{tabular}&
    \begin{tabular}[c]{@{}c@{}} abuse, car accident\\explosion, fight, \\riot, shoot: at \\multiple location \end{tabular}&
     \begin{tabular}[c]{@{}c@{}}violence \\detection~\cite{pang2021violence}, \\anomaly \\detection~\cite{wu2021learning}\end{tabular}&
      \begin{minipage}{.2\textwidth}
      \includegraphics[scale=0.55]{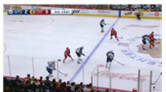}
    \end{minipage}&
    \begin{minipage}{.2\textwidth}
      \includegraphics[scale=0.55]{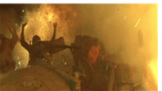}
    \end{minipage}\\ \hline
     
     \begin{tabular}[c]{@{}c@{}} BAREM~\cite{belmonte2021barem}\\(2021)\end{tabular}&
    \begin{tabular}[c]{@{}c@{}}frustration: on \\e-service platform\end{tabular}&
     \begin{tabular}[c]{@{}c@{}}Behaviour Analysis\\ for Reverse Efficient\\ Modeling~\cite{belmonte2021barem}\end{tabular}&
      \begin{minipage}{.2\textwidth}
      \includegraphics[scale=0.2]{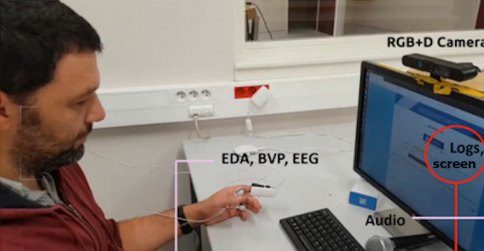}
    \end{minipage}&
    Available upon request \\ \hline
   
\end{tabular}
}
\end{table*}

%%%%%%%%%%%%%%%%%%%%%%%%%%%%%%%%%%%%%%%%%
%========================================
% %-------------------------------
% \begin{figure}[!htbp]
%   \centering
%     \includegraphics[scale=0.6]{DRAWN/audioVisual_taxonomy_labeling_based.png}
%   \caption{Categorization of audio-visual anomaly datasets based on labeling procedure}
%   \label{fig:audioVideo_taxonomy_labeling_based}
% \end{figure}
% %-------------------------------
% \subsection{Labelling procedure}
% Labeling-based categorization of audio-visual datasets is shown via Fig.~\ref{fig:audioVideo_taxonomy_labeling_based}. 
% As it can be seen from the figure, all the datasets possess only frame-level annotations. Therefore, anomaly localization is not yet explored in audio-visual datasets.
%%%%%%%%%%%%%%%%%%%%%%%%%%%%%%%%%%%%%%%%%%%%%%%%%%%%%%%%%%%%%%%
\subsection{Dataset Overview}
% %There are very few audio-visual datasets for anomaly detection. 
For a quick analysis of all the publicly available audio-visual anomaly datasets, we list them in increasing order of their release year via Table~\ref{tab:audioVideoTableMain}. The table has other information, viz., where it was collected, what anomalies are in them, applications where it has been used for bench-marking, a normal sample image, and an abnormal sample image.
These datasets mainly focus on the specific type of anomaly. 
VSD~\cite{demarty2015vsd} dataset, XD-Violence~\cite{wu2020not} dataset, etc., consist of only fight events, whereas EMOLY~\cite{fayet2018emo} dataset has an anomalous mood of a person as an anomaly. Human-human interaction dataset has agitation behavior as an anomaly. Apart from having one class of anomaly, the datasets are mainly recorded in a controlled environment (except for fight anomaly, which is easy to collect) by some human actors. The existing audio-visual dataset is good for application-specific anomaly detection; however, there is a lack of audio-visual datasets for generic scene surveillance. Some researchers have tried collecting such datasets, but they are not released publicly due to privacy or legal issues. 
%%%%%%%%%%%%%%%%%%%%%%%%%%%%%%%%%%%%%%%%%%
\section{Discussion: Towards the future}\label{sec:futureDirection}
During the last decade, there has been a drastic shift from datasets with less number of anomaly samples and total duration to those with diverse range of anomalies and gigantic volume. It may be observed that the availability of datasets with crime-specific anomalies has facilitated the development of automated surveillance frameworks for crime detection. For videos, the category spans detection of the explosion, abuse, panic escape, assault, accident, violence, fight, etc. In the case of audio, they span as detection of the gunshot, shout, etc. Further, for the audio-visual dataset, the categories span as detection of riots, fight, and violence. 

By analyzing the applications and anomalies present in the datasets discussed across Sections~\ref{sec:videoDatasets} to \ref{sec:audioVisual}, we can see that they have mainly specific types of anomalies. There are a very few datasets of heterogeneous nature. This is almost zero in the audio-visual category. Thus, existing datasets are useful for specific anomaly detection, particularly crime-oriented anomaly detection, traffic rule violation, etc. However, for future applications like smart city surveillance we need generic scene monitoring where we have to raise an alarm for interesting events, which may or may not be crime-oriented. This shows a strong need to develop datasets having diverse ranges of anomalous samples. Also, the datasets in the audio-visual category mainly contain acted anomalies, which limits its usefulness. Even if some acted situation/events need to be added, it should be in such a manner that it appears natural and contains the necessary amount of variations mimicking real life.

Apart from this, all the existing multimedia datasets lack anomalies with concept drift~\cite{kumari2020multivariate}. If an event/ object is regarded anomaly in a dataset, it is always regarded as an anomaly for that scene, no matter how frequently it may occur in the distant future. This is due to the fact that datasets are too short to contain this effect. We should pay attention that the datasets which are more than a duration of 5-10 hours are not suitable here because the samples are collected from different time-stamp, location, and have only specific anomalies in rare amounts. Thus, interclass shift, i.e., abnormal to normal class and vice-versa, is not observed. There are attempts to record long untrimmed footage at one place, e.g., QMUL~\cite{loy2008local}, ADOC~\cite{pranav2020day}, etc. However, the authors do not attempt to provide annotations in accordance with concept drift.
%%%%%%%%%%%%%%%%%%%%%%%%%%%%%%%%%%%%%%%%%
\section{Conclusion}\label{sec:conclusions}
This paper presents a survey of multimedia datasets for anomaly detection to researchers working towards automated surveillance. The structured comparison of datasets on various attributes also helps to understand datasets better. There are a large number of short-length and giant video datasets available. Some are developed for generic scene surveillance, whereas other are specific anomaly datasets. Datasets for heterogeneous anomaly are far less compared to specific anomaly datasets. 
In case of audio, datasets are mainly developed for machine surveillance such as defect detection, fault detection, etc. Generally, surveillance using audio alone in outdoor scenarios is not efficient due to the presence of multiple auditory signals superimposed together. However, when analyzed together with video, they can offer crucial and complementary information about the target scene. However, the datasets developed towards audio-visual surveillance are far less compared to that for audio or video. There are a few audio-visual datasets for a specific action, such as fight and agitation detection. A recently released dataset, viz., the EMOLY dataset, has used only the upper body of individuals and their speech information to facilitate abnormal behavior detection. However, there is a strong need to develop more audio-visual datasets for generic scene surveillance. We believe the survey presented in this article will help the prospective researchers who intend to contribute datasets or research in this field.
%%%%%%%%%%%%%%%%%%%%%%%%%%%%%%%%%%%%%%%%%%%%%%%%%%%%%%%%%%%%%%%%%

\bibliographystyle{IEEEtran}
\bibliography{main.bib}
\end{document}